%% file: ms.tex
\documentclass[11pt,a4paper,table,xcdraw]{article}
\pdfoutput=1
\usepackage[hyperref]{acl2019}
\usepackage{times}
\usepackage{latexsym}
\usepackage{multirow}
\usepackage{xcolor}
\usepackage{array,graphicx}
\usepackage{amsmath}
\usepackage{verbatim}
\usepackage{subfig}
\usepackage{caption}

\usepackage{url}
\usepackage{todonotes}

\aclfinalcopy % Uncomment this line for the final submission
\def\aclpaperid{2198} %  Enter the acl Paper ID here

\title{Is Attention Interpretable?}

\author{Sofia Serrano$^\ast$ \quad Noah A. Smith$^{\ast\dagger}$\\
  $^\ast$Paul G. Allen School of Computer Science \& Engineering,\\
  University of Washington,
  Seattle, WA, USA \\
  $^\dagger$Allen Institute for Artificial Intelligence, Seattle, WA, USA \\
  {\tt \{sofias6,nasmith\}@cs.washington.edu} \\}

\date{}

\begin{document}
\maketitle
\begin{abstract}
    Attention mechanisms have recently boosted performance on a range of NLP tasks. Because attention layers explicitly weight input components' representations, it is also often assumed that attention can be used to identify information that models found important (e.g., specific contextualized word tokens). We test whether that assumption holds by manipulating attention weights in already-trained text classification models and analyzing the resulting differences in their predictions. While we observe some ways in which higher attention weights correlate with greater impact on model predictions, we also find many ways in which this does not hold, i.e., where gradient-based rankings of attention weights better predict their effects than their magnitudes. We conclude that while attention noisily predicts input components' overall importance to a model, it is by no means a fail-safe indicator.\footnote{Code is available at \url{https://github.com/serrano-s/attn-tests}.}
\end{abstract}

\input{intro-background}

\input{testing-for-interpretability}

\input{data-models}

\input{single-weights}

\input{sets-of-weights}

\section{Limitations}

There are important limitations to the work we describe here, perhaps the most important of which is our focus on text classification. By choosing to focus on this task, we use the fact that decision flips are often not trivially achieved to ground our judgments of importance in model decision changes. However, for a task with a much larger output space (such as language modeling or machine translation) where almost anything might flip the decision, decision flips are likely too coarse a signal to identify meaningful differences. Determining an analogous informative threshold in changes to model outputs would be key to expanding this sort of analysis to other groups of models.

A related limitation is our reliance in many of these tests on a fairly strict definition of importance tied to the output's argmax; an alternative definition of importance might assert that the highest attention weights should identify the most influential representations in pushing towards \textit{any} output class, not just the argmax. Two of the core challenges that would need to be solved to test for how well attention meets this relaxed criterion would be meaningfully evaluating a single attended item's ``importance'' to multiple output classes for comparison to other attended items and, once again, determining what would truly indicate being ``most influential'' in the absence of decision flips as a guide to the output space.

Also, while we explore several model architectures in this work, there exist other attention functions such as multi-headed and scaled dot-product \citep{transformer}, as well as cases where a single attention layer is responsible for producing more than one attended representation, such as in self-attention \cite{cheng2016long}. These variants could have different interpretability properties. Likewise, we only evaluate on final layers of attention here; in large models, lower-level layers of attention might learn to work differently.

\section{Related and Future Work}

We have adopted an erasure-based approach to probing the interpretability of computed attention weights, but there are many other possible approaches. For example, recent work has focused on which training instances \cite{influencefns} or which human-interpretable features were most relevant for a particular decision \cite{lime, arras2016explaining}. Others have explored alternative ways of comparing the behavior of proposed explanation methods \cite{sanitychecks}. Yet another line of work focuses on aligning models with human feedback for what is interpretable \cite{fyshe2015compositional, spine}, which could refine our idea of what defines a high-quality explanation derived from attention.

Finally, another direction for future work would be to extend the importance-ranking comparisons that we deploy here for evaluation purposes into a method for deriving better, more informative rankings, which in turn could be useful for the development of new, more interpretable models.

\section{Conclusion}

It is frequently assumed that attention is a tool for interpreting a model, but we find that attention does not necessarily correspond to importance. In some ways, the two correlate: comparing the highest attention weight to a lower weight, the high attention weight's impact on the model is often larger. However, the picture becomes bleaker when we consider the many cases where the highest attention weights fail to have a large impact.
Examining these cases through multi-weight tests, we see that attention weights often fail to identify the sets of representations most important to the model's final decision. Even in cases when an attention-based importance ranking flips the model's decision faster than an alternative ranking, the number of zeroed attended items is often too large to be helpful as an explanation. We also see a marked effect of the contextualization scope preceding the attention layer on the number of attended items affecting the model's decision; while attention magnitudes do seem more helpful in uncontextualized cases, their lagging performance in retrieving decision rationales elsewhere is cause for concern. What is clear is that in the settings we have examined, attention is not an optimal method of identifying which attended elements are responsible for an output. Attention may yet be interpretable in other ways, but as an importance ranking, it fails to explain model decisions.

\section*{Acknowledgments}

This research was supported in part by a grant from the Allstate Corporation; findings do not necessarily represent the views of the sponsor.
We thank R. Andrew Kreek, Paul Koester, Kourtney Traina, and Rebecca Jones for early conversations leading to this work. We also thank Omer Levy, Jesse Dodge, Sarthak Jain,  Byron Wallace, and Dan Weld for helpful conversations, and our anonymous reviewers for their feedback.

\bibliography{ms}
\bibliographystyle{acl_natbib}

\clearpage\newpage

\appendix

\input{appendices}

\end{document}

%% file: intro-background.tex
\section{Introduction}

Interpretability is a pressing concern for many current NLP models. As they become increasingly complex and learn decision-making functions from data, ensuring our ability to understand why a particular decision occurred is critical.

Part of that development has been the incorporation of attention mechanisms \cite{bahdanau} into models for a variety of tasks. For many different problems---to name a few, machine translation \cite{luong2015effective}, syntactic  parsing \cite{vinyals2015grammar}, reading comprehension \cite{hermann2015teaching}, and language modeling \cite{liu2018learning}---incorporating attention mechanisms into models has proven beneficial for performance. While there are many variants of attention \cite{transformer}, each formulation consists of the same high-level goal: calculating nonnegative weights for each input component (e.g., word) that together sum to 1, multiplying those weights by their corresponding representations, and summing the resulting vectors into a single fixed-length representation.

Since attention calculates a distribution over inputs, prior work has used attention as a tool for interpretation of model decisions \cite{wang2016attention, lee2017interactive, 2dsentemb, ghaeini2018interpreting}. 
The existence of so much work on visualizing attention weights is a testament to attention's popularity in this regard; to name just a few examples of these weights being examined to understand a model, recent work has focused on goals from explaining and debugging the current system's decision \cite{lee2017interactive,ding2017visualizing} to distilling important traits of a dataset \cite{yang2017satirical,beforenamecalling}.

Despite this, existing work on interpretability is only beginning to assess what computed attention weights actually communicate. In an independent and contemporaneous study, \citet{attentionexplanation} explore whether attention mechanisms can identify the relative importance of inputs to the full model, finding them to be highly inconsistent predictors. In this work, we apply a different analysis based on \emph{intermediate} representation erasure to assess whether attention weights can instead be relied upon to explain the relative importance of the inputs to the attention layer itself. We find similar cause for concern: attention weights are only noisy predictors of even intermediate components' importance, and should not be treated as justification for a decision.

%% file: testing-for-interpretability.tex
\section{Testing for Informative Interpretability}

We focus on five- and ten-class text classification models incorporating attention, as explaining the reasons for text classification has been a particular area of interest for recent work in interpretability \cite{han, lime, lei2016rationalizing, pathologies}. 

In order for a model to be interpretable, it must not only suggest explanations that make sense to people, but also ensure that those explanations accurately represent the true reasons for the model's decision.  Note that this type of analysis does not rely on the true labels of the data; if a model produces an incorrect output, but a faithful explanation for which factors were important in that calculation, we still consider it interpretable.

We take the implied explanation provided by visualizing attention weights to be a ranking of importance of the attention layer's input representations, which we denote $\mathcal{I}$:  if the attention allocated to item $i \in \mathcal{I}$ is higher than that allocated to item $j \in \mathcal{I}$, then $i$ is presumed ``more important'' than $j$ to the model's output. In this work, we focus on  whether the attention weights' suggested importance ranking of $\mathcal{I}$ faithfully describes why the model produced its output, echoing existing work on explanation brittleness for other model components \cite{ghorbani2017interpretation}.

\subsection{Intermediate Representation Erasure}

\begin{figure}
	\includegraphics[width=\linewidth]{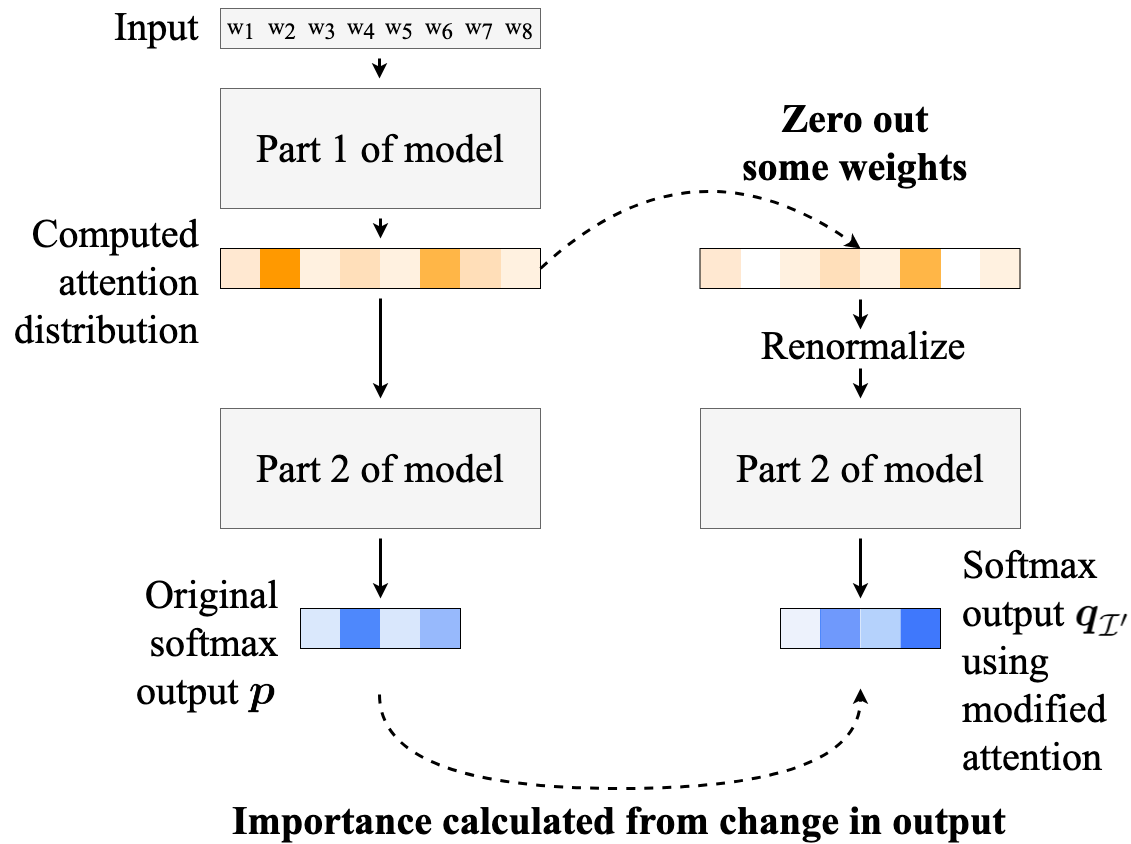}
    \caption{Our method for calculating the importance of representations corresponding to zeroed-out attention weights, in a hypothetical setting with four output classes .}
    \label{defining-importance}
\end{figure}

We are interested in the impact of some contextualized inputs to an attention layer, $\mathcal{I}' \subset \mathcal{I}$, on the model's output. To examine the importance of $\mathcal{I}'$, we run the classification layer of the model twice (Figure~\ref{defining-importance}): once without any modification, and once after renormalizing the attention distribution with $\mathcal{I}'$'s attention weights zeroed out, similar to other erasure-based work \cite{li2016understanding,pathologies}. We then observe the resulting effects on the model's output. We erase at the attention layer to isolate the effects of the attention layer from the encoder preceding it. Our reasoning behind renormalizing is to keep the output document representation from artificially shrinking closer to $\boldsymbol{0}$ in a way never encountered during training, which could make subsequent measurements unrepresentative of the model's behavior in spaces to which it \textit{does} map inputs.

One point worth noting is the facet of interpretability that our tests are designed to capture. By examining only how well attention represents the importance of intermediate quantities, which may themselves already have changed uninterpretably from the model's inputs, we are testing for a relatively low level of interpretability. So far, other work looking at attention has examined whether attention suffices as a holistic explanation for a model's decision \cite{attentionexplanation}, which is a higher bar. We instead focus on the lowest standard of interpretability that attention might be expected to meet, ignoring prior model layers.

We denote the output distributions (over labels) as $\boldsymbol{p}$ (the original) and $\boldsymbol{q}_{\mathcal{I}'}$ (where we erase attention for $\mathcal{I}'$).  The question now becomes how to operationalize ``importance'' given $\boldsymbol{p}$ and $\boldsymbol{q}_{\mathcal{I}'}$. There are many quantities that could arguably capture information about importance. We focus on two: the Jensen-Shannon (JS) divergence between output distributions $\boldsymbol{p}$ and $\boldsymbol{q}_{\mathcal{I}'}$, and whether the argmaxes of $\boldsymbol{p}$ and $\boldsymbol{q}_{\mathcal{I}'}$ differ, indicating a decision flip.

%% file: data-models.tex
\section{Data and Models}

\input{datasets}

We investigate four  model architectures on a topic classification dataset (Yahoo Answers; \citealp{yahooamazon}) and on three review ratings datasets: IMDB \cite{imdb},\footnote{downloaded from \url{github.com/nihalb/JMARS}} Yelp 2017,\footnote{from \url{www.yelp.com/dataset_challenge}} and Amazon \cite{yahooamazon}.
Statistics for each dataset are listed in Table~\ref{dataset-stats}.

\begin{figure}
	\includegraphics[width=\linewidth]{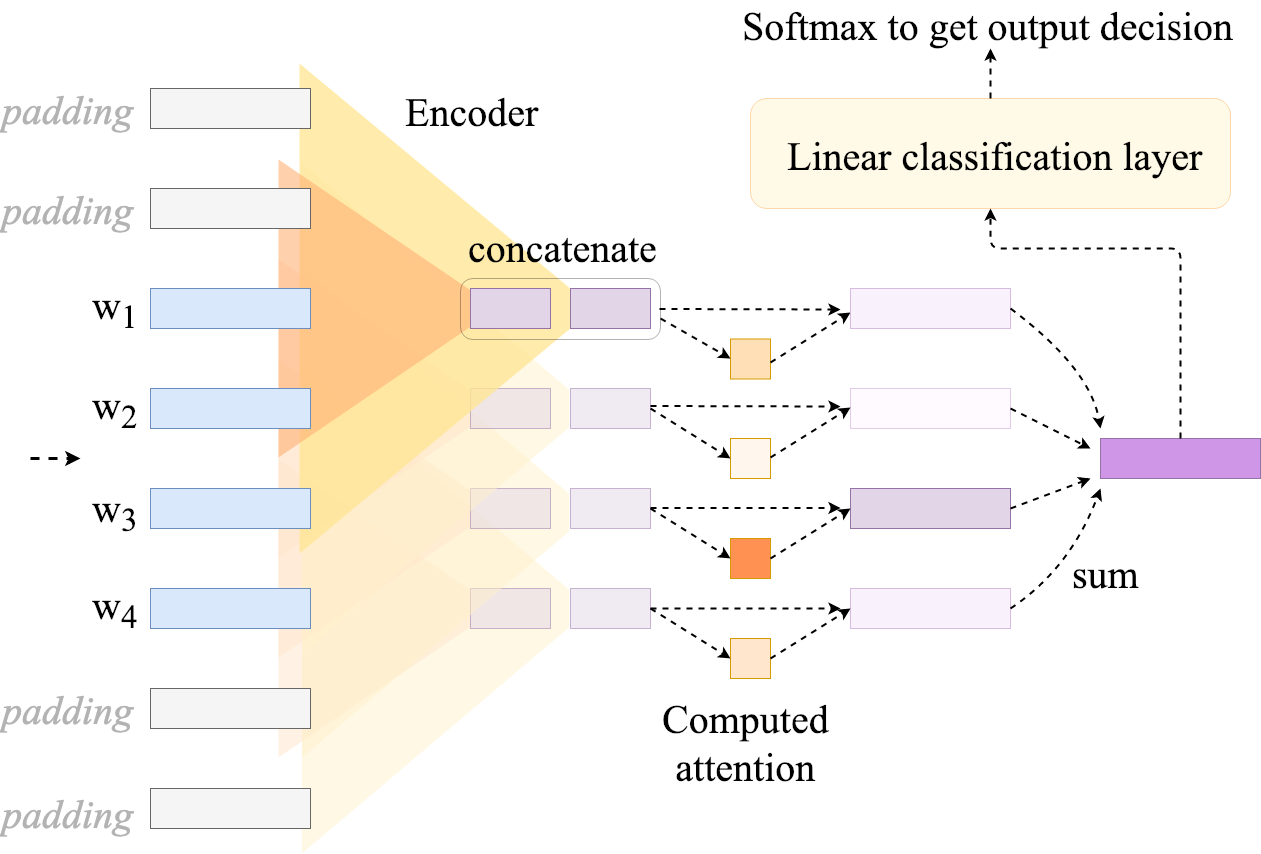}
    \caption{Flat attention network (FLAN) demonstrating a convolutional encoder. Each contextualized word representation is the concatenation of two sizes of convolutions: one applied over the input representation and its two neighbors to either side, and the other applied over the input representation and its single neighbor to either side. For details, see Appendix A.1.}
    \label{model-architecture}
\end{figure}

Our model architectures are inspired by the hierarchical attention network \citep[HAN;][]{han}, a text classification model with two layers of attention, first to the word tokens in each sentence and then to the resulting sentence representations. The layer that classifies the document representation is linear with a softmax at the end. 

We conduct our tests on the softmax formulation of attention,\footnote{Alternatives such as sparse attention \citep{sparsemax} and unnormalized attention \citep{yangfeng-acl-2017} have been proposed.} which is used by most models, including the HAN.
Specifically, we use the additive formulation originally defined in \citet{bahdanau}. Given attention layer $\ell$'s learned parameters, element $i$ of a sequence, and its encoded representation $\mathbf{h}_i$, the attention weight $\alpha_i$ is computed using $\ell$'s learned context vector $\mathbf{c_\ell}$ as follows:
\begin{align*}
    \mathbf{u}_{i} =& \tanh(\mathbf{W}_\ell \mathbf{h}_{i} + \mathbf{b}_\ell) \\
    \alpha_{i} =& \frac{\exp{\mathbf{u}_{i}^\top \mathbf{c}_\ell}}{\sum_i{\exp{\mathbf{u}_{i}^\top \mathbf{c}_\ell}}} 
\end{align*}
We evaluate on the original HAN architecture, but we also vary that architecture in two key ways:
\begin{enumerate}
\item{Number of attention layers: besides exploring models with a final layer of attention over sentence representations (which we denote with a ``HAN'' prefix), we also train ``flat'' attention networks with only one layer of attention over all contextualized word tokens (which we denote with a ``FLAN'' prefix). In either case, though, we only run tests on models' final layer of attention.}
\item{Reach of encoder contextualization: The original HAN uses recurrent encoders to contextualize input tokens prior to an attention layer (specifically, bidirectional GRUs running over the full sequence). Aside from biRNNs, we also experiment with models that instead contextualize word vectors by convolutions on only a token's close neighbors, inspired by \citet{kim2014convolutional}. See Figure~\ref{model-architecture} for a diagram of the FLAN architecture using a convolutional encoder. We denote this variant of an architecture with a ``conv'' suffix. Finally, we also test models that are trained with no contextualizing encoder at all; we denote these with a ``noenc'' suffix.}
\end{enumerate}
The classification accuracy of each of our trained models is listed in Table~\ref{model-accuracy} in the appendix, along with training details for the different models.

%% file: datasets.tex
\begin{table*}[t!]
\begin{center}
\begin{tabular}{lrrrrrrr}
\hline \bf Dataset & \hspace{-.35in} \bf Av.~\# Words & \small\bf (s.d.) &  \bf Av.~\# Sents. & \small\bf (s.d.) & \bf \# Train.~+ Dev. & \bf \# Test & \bf \# Classes \\ \hline
Yahoo Answers & 104 &\small (114) & 6.2 &\small (5.9) & 1,400,000 & 50,000 & 10\\
IMDB & 395 &\small (259) & 16.2 &\small (10.7) & 122,121 & 13,548 & 10\\
Amazon & 73 &\small (48) & 4.3 &\small (2.6) & 3,000,000 & 650,000 & 5\\
Yelp & 125 &\small (109) & 7.0 &\small (5.6) & 650,000 & 50,000 & 5\\
\hline
\end{tabular}
\end{center}
\caption{\label{dataset-stats} Dataset statistics.}
\end{table*}

%% file: single-weights.tex
\section{Single Attention Weights' Importance}

As a starting point for our tests, we investigate the relative importance of attention weights when only one weight is removed. Let $i^\ast \in \mathcal{I}$ be the component with the highest attention and let $\alpha_{i^\ast}$ be its attention. We compare $i^\ast$'s importance to some other attended item's importance in two ways.

\subsection{JS Divergence of Model Output Distributions} \label{js}

\begin{figure}[!htbp]
	\includegraphics[width=\linewidth]{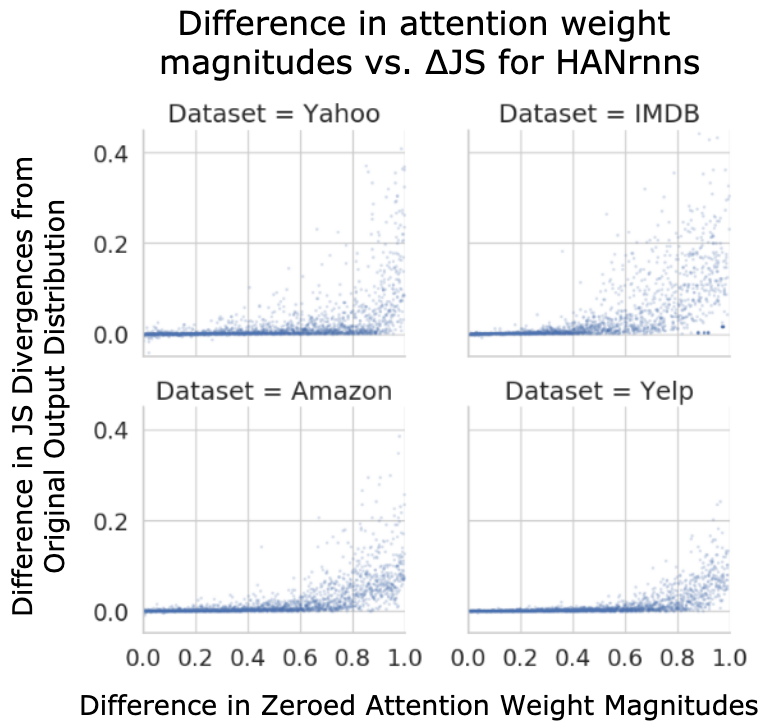}
    \caption{Difference in attention weight magnitudes versus $\Delta\mathrm{JS}$ for HANrnns, comparable to results for the other architectures; for their plots, see Appendix A.2.}
    \label{differences_in_jsdivs_vs2nd_hanrnn}
\end{figure}

We wish to compare how $i^\ast$'s impact on the model's output distribution compares to the impact corresponding to a random attended item $r$ drawn uniformly from $\mathcal{I}$. Our first approach to this will be to calculate two JS divergences---one being the JS divergence of the model's original \textit{output} distribution from its output distribution after removing only $i^\ast$, and the other after removing only $r$---and compare them to each other. We subtract the output JS divergence after removing $r$ from the output JS divergence after removing $i^\ast$:
\begin{equation}
  \Delta\mathrm{JS} =  \mathrm{JS}(\boldsymbol{p}, \boldsymbol{q}_{\{i^\ast\}}) - \mathrm{JS}(\boldsymbol{p}, \boldsymbol{q}_{\{r\}}) \label{js-delta}
\end{equation}
We plot this quantity against the difference $\Delta\alpha = \alpha_{i^\ast} - \alpha_{r}$ in Figure~\ref{differences_in_jsdivs_vs2nd_hanrnn}.
We show results on the HANrnn, as the trends for the other models are very similar; see Figures~\ref{all-delta-js}--\ref{all-delta-js-hists} and the tables in Figure~\ref{dec-flip-tables-attn} in the Appendix for full results.

\begin{figure}[!htbp]
	\includegraphics[width=\linewidth]{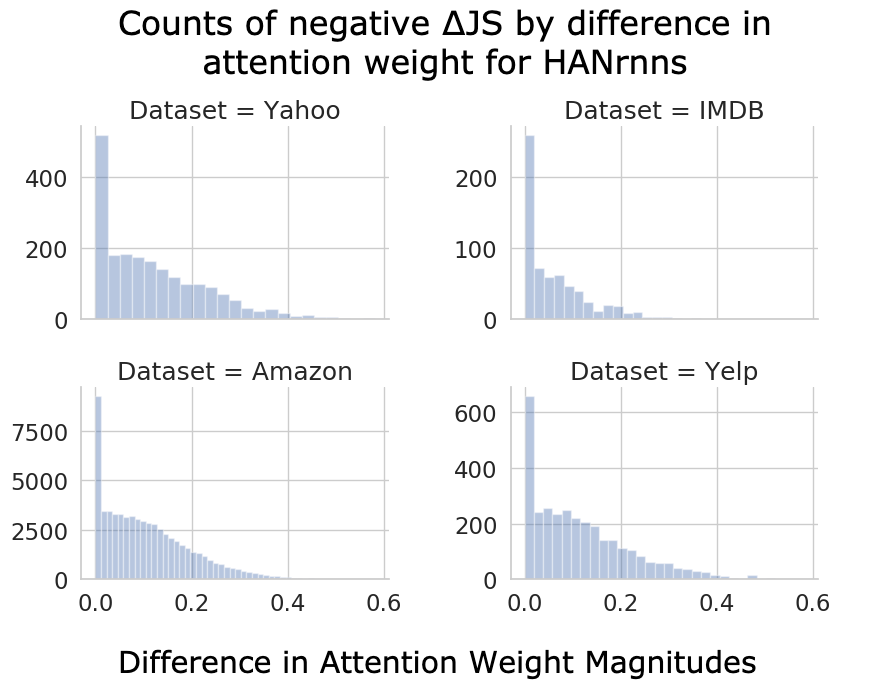}
    \caption{These are the counts of test instances for the HANrnn models for which $i^\ast$'s JS divergence was smaller, binned by $\Delta\alpha$. These counts comprise a small fraction of the test set sizes listed in Table~\ref{dataset-stats}.}
    \label{differences_in_jsdivs_hanrnn}
\end{figure}

Intuitively, if $i^\ast$ is truly the most important, then we would expect Eq.~\ref{js-delta} to be positive, and that is what we find the vast majority of the time. In addition, examining Figure~\ref{differences_in_jsdivs_vs2nd_hanrnn}, we see that nearly all negative $\Delta\mathrm{JS}$ values are close to 0. By binning occurrences of negative $\Delta\mathrm{JS}$ values by the difference between $\alpha_{i^\ast}$ and $\alpha_r$ in Figure~\ref{differences_in_jsdivs_hanrnn}, we also see that in the cases where $i^\ast$ had a smaller effect, the gap between $i^\ast$'s attention and $r$'s tends to be small. This is encouraging, indicating that in these cases, $i^\ast$ and $r$ are nearly ``tied'' in attention. 

However, the picture of attention's interpretability grows somewhat more murky when we begin to consider the magnitudes of positive $\Delta\mathrm{JS}$ values in Figure~\ref{differences_in_jsdivs_vs2nd_hanrnn}. We notice across datasets that even for quite large differences in attention weights like 0.4, many of the positive $\Delta\mathrm{JS}$ are still quite close to zero.
Although we do finally see an upward swing in $\Delta\mathrm{JS}$ values once $\Delta\alpha$ gets even larger, indicating only one very high attention weight in the distribution, this still leaves many open questions about exactly how much difference in impact $i^\ast$ and $r$ can typically be expected to have.

\subsection{Decision Flips Caused by Zeroing Attention}
\input{decflip_rand}
Since attention weights are often interpreted as an explanation for a model's argmax decision, our second test looks at another, more immediately visible change in model outputs: decision flips. 
For clarity, we limit our discussion to results for the HANrnns, which reflect the same patterns observed for the other architectures. (Results for all other models are in Appendix A.2.)

Table~\ref{rand-decflip-table} shows, for each dataset, a contingency table for the two binary random variables (i) does zeroing $\alpha_{i^\ast}$ (and renormalizing) result in a decision flip? and (ii) does doing the same for a different randomly chosen weight $\alpha_r$ result in a decision flip?  To assess the comparative importance of $i^\ast$ and $r$, we consider when exactly one erasure changes the decision (off-diagonal cells).  For attention to be interpretable, the blue, upper-right values ($i^\ast$, not $r$, flips a decision) should be much larger than the orange, lower-left values ($r$, not $i^\ast$, flips a decision), which should be close to zero.\footnote{We see this pattern especially strongly for FLANs (see Appendix), which is unsurprising since $\mathcal{I}$ is all \emph{words} in the input text, so most attention weights are very small.}

  Although for some datasets in Table~\ref{rand-decflip-table}, the ``orange'' values are non-negligible, we mostly see that their fraction of total off-diagonal values mirrors the fraction of negative occurrences of Eq. 1 in Figure \ref{differences_in_jsdivs_hanrnn}.
  However, it's somewhat startling that in the vast majority of cases, erasing $i^\ast$ does \emph{not} change the decision (``no'' row of each table).  This is likely explained in part by the signal pertinent to the classification being distributed across a document (e.g., a ``Sports'' question in the Yahoo Answers dataset could signal ``sports'' in a few sentences, any one of which suffices to correctly categorize it). However, given that these results are for the HAN models, which typically compute attention over ten or fewer sentences, this is surprising.

Altogether, examining importance from a single-weight angle paints a tentatively positive picture of attention's interpretability, but also raises several questions about the many cases where the difference in impacts between $i^\ast$ and $r$ is almost identical (i.e., $\Delta\mathrm{JS}$ values close to 0 or the many cases where neither $i^\ast$ nor $r$ cause a decision flip). To answer these questions, we require tests with a broader scope.

%% file: decflip_rand.tex
\begin{table}[!htbp]
\begin{tabular}{clccllcc}
\multicolumn{1}{l}{}                &      & \multicolumn{6}{c}{\textbf{Remove random: Decision flip?}}                                                                                                                                                                                                                  \\
                                          &                           & \multicolumn{2}{c}{Yahoo}                                                                       &  &                           & \multicolumn{2}{c}{IMDB}                                                                        \\
                                          &                           & Yes                                           & No                                           &  &                           & Yes                                           & No                                           \\ \cline{3-4} \cline{7-8} 
                                          & \multicolumn{1}{r|}{Yes} & \multicolumn{1}{c|}{0.5}                         & \multicolumn{1}{c|}{\cellcolor[HTML]{DAE8FC}8.7} &  & \multicolumn{1}{r|}{Yes} & \multicolumn{1}{c|}{2.2}                         & \multicolumn{1}{c|}{\cellcolor[HTML]{DAE8FC}12.2} \\ \cline{3-4} \cline{7-8} 
                                          & \multicolumn{1}{r|}{No} & \multicolumn{1}{c|}{\cellcolor[HTML]{F8A102}1.3} & \multicolumn{1}{c|}{89.6}                         &  & \multicolumn{1}{r|}{No} & \multicolumn{1}{c|}{\cellcolor[HTML]{F8A102}1.4} & \multicolumn{1}{c|}{84.2}                         \\ \cline{3-4} \cline{7-8} 
                                          &                           & \multicolumn{1}{l}{}                           & \multicolumn{1}{l}{}                           &  &                           & \multicolumn{1}{l}{}                           & \multicolumn{1}{l}{}                           \\
                                          &                           & \multicolumn{2}{c}{Amazon}                                                                      &  &                           & \multicolumn{2}{c}{Yelp}                                                                        \\
                                          &                           & Yes                                           & No                                           &  &                           & Yes                                           & No                                           \\ \cline{3-4} \cline{7-8} 
                                          & \multicolumn{1}{r|}{Yes} & \multicolumn{1}{c|}{2.7}                         & \multicolumn{1}{c|}{\cellcolor[HTML]{DAE8FC}7.6} &  & \multicolumn{1}{r|}{Yes} & \multicolumn{1}{c|}{1.5}                         & \multicolumn{1}{c|}{\cellcolor[HTML]{DAE8FC}8.9} \\ \cline{3-4} \cline{7-8} 
\multirow{-9}{*}{\rotatebox[origin=c]{90}{\textbf{Remove $i^\ast$: Decision flip?}}} & \multicolumn{1}{r|}{No} & \multicolumn{1}{c|}{\cellcolor[HTML]{F8A102}2.7} & \multicolumn{1}{c|}{87.1}                         &  & \multicolumn{1}{r|}{No} & \multicolumn{1}{c|}{\cellcolor[HTML]{F8A102}1.9} & \multicolumn{1}{c|}{87.7}                         \\ \cline{3-4} \cline{7-8} 
\end{tabular}
\caption{\label{rand-decflip-table} Percent of test instances in each decision-flip indicator variable category for each HANrnn.}
\end{table}

%% file: sets-of-weights.tex
\section{Importance of Sets of Attention Weights}

Often, we care about determining the \emph{collective} importance of a set of components $\mathcal{I}'$. To address that aspect of attention's interpretability and close gaps left by single-weight tests, we introduce tests to determine how multiple attention weights perform together as importance predictors.

\subsection{Multi-Weight Tests}

For a hypothesized ranking of importance, such as that implied by attention weights, we would expect the items at the top of that ranking to function as a concise explanation for the model's decision. The less concise these explanations get, and the farther down the ranking that the items truly driving the model's decision fall, the less likely it becomes for that ranking to truly describe importance. In other words, we expect that the top items in a truly useful ranking of importance would comprise a minimal necessary set of information for making the model's decision.

The idea of a minimal set of inputs necessary to uphold a decision is not new;  \citet{li2016understanding}  use reinforcement learning to attempt to construct such a minimal set of words, \citet{lei2016rationalizing} train an encoder to constrain the input prior to classification, and much of the work that has been done on extractive summarization takes this concept as a starting point \citep{lin2011class}. However, such work has focused on approximating minimal sets, instead of evaluating the ability of other importance-determining ``shortcuts'' (such as attention weight orderings) to identify them. \citet{nguyen} leveraged the idea of minimal sets in a much more similar way to our work, comparing different input importance orderings.

Concretely, to assess the validity of an importance ranking method (e.g., attention), we begin erasing representations from the top of the ranking downward until the model's decision changes. Ideally, we would then enumerate all possible subsets of that instance's components, observe whether the model's decision changed in response to removing each subset, and then report whether the size of the minimal decision-flipping subset was equal to the number of items that had needed to be removed to achieve a decision flip by following the ranking. However, the exponential number of subsets for any given instance's sequence of components (word or sentence representations, in our case) makes such a strategy computationally prohibitive, and so we adopt a different approach. 

Instead, in addition to our hypothesized importance ranking (attention weights), we consider alternative rankings of importance; if, using those, we repeatedly discover cases where removing a smaller subset of items would have sufficed to change the decision, this signals that our candidate ranking is a poor indicator of importance.

\subsection{Alternative Importance Rankings}

Exhaustively searching the space of component subsets would be far too time-consuming in practice, so we introduce three other ranking schemes.

The first is to randomly rank importance. We expect that this ranking will perform quite poorly, but it provides a point of comparison by which to validate that ranking by descending attention weights is at least somewhat informative.

The second ranking scheme, inspired by \citet{compositionality} and \citet{pathologies}, is to order the attention weights by the gradient of the decision function with respect to each calculated attention weight, in descending order. Since each of the datasets on which we evaluate has either five or ten output classes, we take the decision function given a real-valued model output vector to be
$$d(\boldsymbol{x}) = \frac{\exp{(\max_i{(\boldsymbol{x}_i}}))}{\sum_i{\exp{\boldsymbol{x}_i}}}.$$

Unlike the last two proposed rankings, our third ranking scheme uses attention weights, but supplements them with information about the gradient. For this ranking, we multiply each of our calculated gradients from our previous proposed ranking scheme by their corresponding attention weight magnitude. Under this ordering, attended items that have both a high attention weight and a high calculated gradient with respect to their attention weight will be ranked most important.

We introduce these last two rankings as an attempt to discover smaller sets not produced by the attention weight ranking. Note, however, that we still do not take either as a gold-standard indicator of importance to the model, as with the gradient in \citet{rightreasons} and \citet{melis2018towards}, but merely as an alternative ordering method. The ``gold standard'' in our case would be the minimal set of attention weights to zero out for the decision to change, which none of our ordering methods will necessarily find for a particular instance.

\subsection{Instances Excluded from Analysis}

In cases where removing all but one input to the attention layer still does not produce a decision flip, we finish the process of removing components by removing the final representation and replacing the output of the attention layer with an arbitrary vector; we use the zero vector for our tests. Even so, since every real-valued vector output by the attention layer is mapped to an output distribution, removing this final item will still not change the classification decision for instances that the model happened to originally map to that same class. We exclude such instances for which the decision never changed from all subsequent analyses.

We also set aside any test instances with a sequence length of 1 for their final attention layer, as there is only one possible ordering for such cases.

\begin{figure*}[!ht]
  \includegraphics[width=\textwidth]{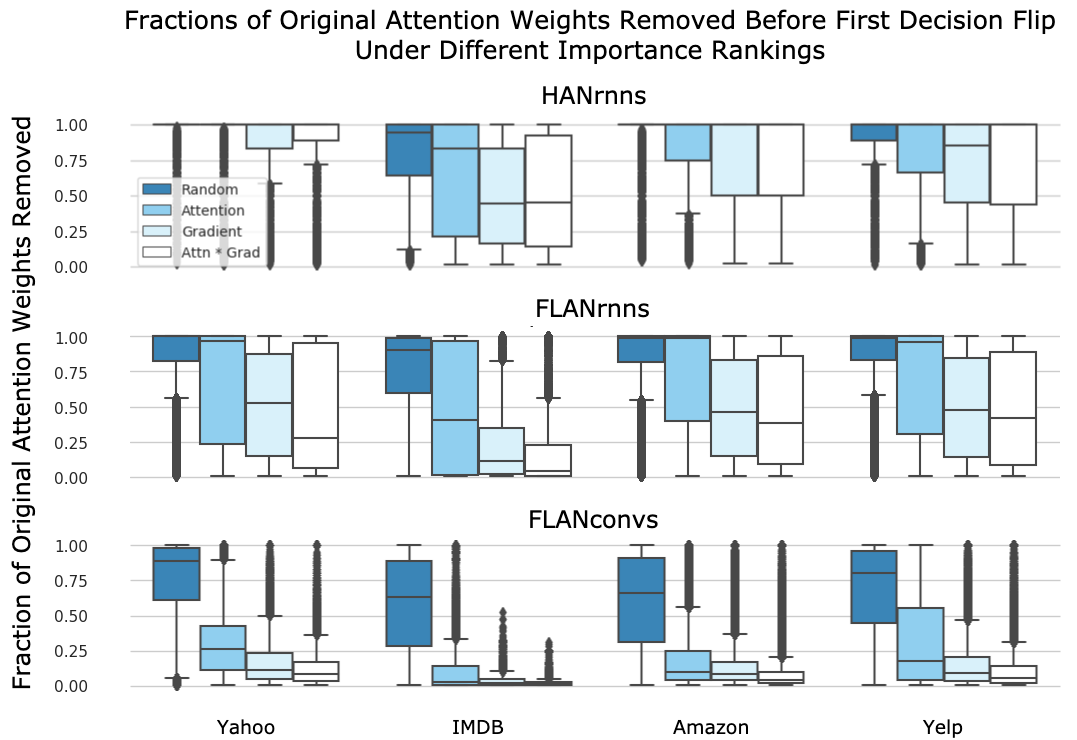}
  \caption{The distribution of fractions of items removed before first decision flips on three model architectures under different ranking schemes. Boxplot whiskers represent the highest/lowest data point within 1.5 IQR of the higher/lower quartile, and dataset names at the bottom apply to their whole column. In several of the plots, the median or lower quartile aren't visible; in these cases, the median/lower quartile is either 1 or very close to 1.}
  \label{fracremoved-boxplots}
\end{figure*} 

\subsection{Attention Does Not Optimally Describe Model Decisions}

Examining our results in Figure~\ref{fracremoved-boxplots}, we immediately see that ranking importance by descending attention weights is not optimal for our models with encoders. While removing intermediate representations in decreasing order by attention weights often leads to a decision flip faster than a random ranking, it also clearly falls short of matching (or even approaching) the decision-flipping efficiency of either the gradient ordering or gradient-attention-product ordering in many cases.

In addition, although the product-based ranking often (but not always) requires slightly fewer removed items than the gradient ranking, we see that the purely gradient-based ranking ignoring attention magnitudes comes quite close to it, far outperforming purely attention-based orderings. For ten of our 16 models with encoders, removing by gradient found a smaller decision-flipping set of items than attention for over 50\% of instances in that model's test set, with that percentage often being much higher. In fact, for \textit{every} model with an encoder that we tested, there were at least 1.6 times as many test instances where the purely gradient-based ranking managed a decision flip faster than the attention-based ranking than vice versa.

We do not claim that ranking importance by either descending gradients or descending gradient-attention products is optimal, but in many cases they discover much smaller decision-flipping sets of items than attention weights. Therefore, ranking representations in descending order by attention weight clearly fails to uncover a minimal set of decision-flipping information much of the time, which is a warning sign that we should be skeptical of trusting groups of attention weight magnitudes as importance indicators.

\subsection{Decision Flips Often Occur Late}

For all ordering schemes we tried, we were struck by the large fraction of items that had to be removed to achieve a decision flip in many models. This is slightly less surprising for the HANs, as they compute attention over shorter sequences of sentences (see Table~\ref{dataset-stats}). For the FLAN models, though, this result is highly unexpected. The sequences across which FLANs compute attention are usually hundreds of tokens in length, meaning most attention weights will likely be minuscule. 

The distributions of tokens removed by our different orderings that we see for the FLANrnns in Figure~\ref{fracremoved-boxplots} are therefore remarkably high, especially given that all of our classification tasks have at least five output classes. We also note that due to the exponential nature of the softmax, softmax attention distributions typically contain only a few high-weighted items before the calculated weights become quite small, which can be misleading. In many cases, flipping the model's original decision requires digging deep into the small attention weights, with the high-weighted components not actually being the reason for the decision.

For several of our models, especially the FLANs (which typically compute attention over hundreds of tokens), this fact is concerning from an explainability perspective. Lipton \citeyearpar{mythos} describes a model as ``transparent'' if ``a person can contemplate the entire model at once.'' Applying this insight to the explanations suggested by attention, if an explanation rests on simultaneously considering hundreds of attended tokens necessary for a decision-- even if that set were minimal---that would still raise serious transparency concerns.

\begin{figure*}[ht]
  \includegraphics[width=\textwidth]{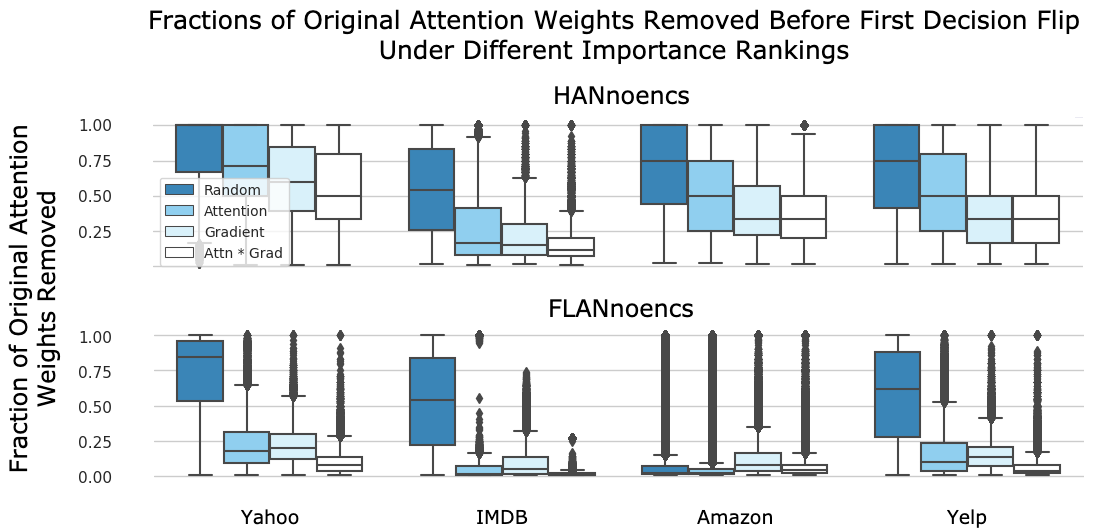}
  \caption{The distribution of fractions of items removed before decision flips on the encoderless model architectures under different ranking schemes. The Amazon FLANnoenc results have a very long tail; using the legend's order of rankings, the percentage of test instances with a removed fraction above 0.50 for that model is 12.4\%, 2.8\%, 0.9\%, and 0.5\%, respectively.}
  \label{fracremoved-enclessboxplots}
\end{figure*}

\subsection{Effects of Contextualization Scope on Attention's Interpretability}

One last question we consider is whether the large number of items that are removed before decision flips can be explained in part by the scope of each model's contextualization. In machine translation, prior work has observed that recurrent encoders over a full sequence can ``shift'' tokens' signal in ways that cause subsequent attention layers to compute unintuitive off-by-one alignments \cite{contextualattn}. We hypothesize that in our text classification setting, the bidirectional recurrent structure of the HANrnn and FLANrnn encoders might instead be redistributing operative signal from a few informative input tokens across many others' contextualized representations. 

Comparing the decision flip results for the FLANconvs in Figure~\ref{fracremoved-boxplots} to those for the FLANrnns supports this theory. We notice decision flips happening much faster than for either of the RNN-based model architectures, indicating that the biRNN effectively does learn to widely redistribute the classification signal. In contrast, the convolutional encoders only allow contextualization with respect to an input token's two neighbors to either side. We see similar results when comparing the two HAN architectures, albeit much more weakly (see Figure~\ref{app-fracremoved} in Appendix A.2); this is likely due to the smaller number of tokens being contextualized by the HANs (sentence representations instead of words), so that contextualization with respect to a token's close neighbors encompasses a much larger fraction of the full sequence.

We see this difference even more strongly when we compare to the encoderless model architectures, as shown in Figure \ref{fracremoved-enclessboxplots}. Compared to both other model architectures, we see the fraction of necessary items to erase for flipping the decision plummet. We also see random orderings mostly do better than before, indicating more brittle decision boundaries, \textit{especially} on the Amazon dataset.\footnote{This is likely due to the fact that with no contextualization, the final attended representations are just a linear combination of the input embeddings, so the embeddings themselves are responsible for learning to directly encode a decision. Since Amazon has the largest ratio of documents (which probably vary in their label) to unique word embeddings by a factor of more than two times any other dataset's, and the final attended representations in the FLANnoencs are unaggregated word embeddings, it stands to reason that the lack of encoders would be a much bigger obstacle in its case.}
In this situation, we see attention magnitudes generally indicate importance on par with (or better than) gradients, but that the product-based ordering is still often a more efficient explanation.

While these differences  themselves are not an argument against attention's interpretability, they highlight the distinction between attention's weighting of intermediate, contextualized representations and the model's use of the original input tokens themselves. Our RNN-based models' ability to maintain their original decision well past the point at which models using only local or no context have lost the signal driving their original decisions confirms that attention weights for a contextualized representation do not necessarily map neatly back to the original tokens. This might at least partly explain the striking near-indifference of the model's decision to the contributions of particular contextualized representations in both our RNN-based models and in \citet{attentionexplanation}, who also use recurrent encoders. 

However, the results from almost all models continue to support that ranking importance by attention is still not optimal; our non-random alternative rankings still uncover many cases where fewer items could be removed to achieve a decision flip than the attention weights imply.

%% file: appendices.tex
\section{Appendices}

\subsection{Model Hyperparameters and Performance}

\input{appendix_stuff/model_performance_table}

We lowercased all tokens during preprocessing and used all hyperparameters specified in \cite{han}, except for those related to the optimization algorithm or, in the case of the convolutional or no-encoder models, the encoder. For each convolutional encoder, we trained two convolutions: one sweeping over five tokens, and one sweeping over three. As the output representation of token $x$, we then concatenated the outputs of the five-token and three-token convolutions centered on $x$. Unless otherwise noted, to train each model, we used Adam \cite{kingma2014adam} with gradient clipping of 10.0 and a patience value of 5, so we would stop training a model if five epochs elapsed without any improvement in validation set accuracy. In addition, for each model, we specified a learning rate for training, and dropout before each encoder layer (or attention layer, for the encoderless models) and also within the classification layer. For the HAN models, these are the values we used:

\begin{itemize}
    \item Yahoo Answers HANrnn, Yahoo Answers HANconv
    \begin{itemize}
        \item Pre-sentence-encoder dropout: 0.4445
        \item Pre-document-encoder dropout: 0.2202
        \item Classification layer dropout: 0.3749
        \item Learning rate: 0.0004
    \end{itemize}
    \item IMDB HANrnn
    \begin{itemize}
        \item Pre-sentence-encoder dropout: 0.4445
        \item Pre-document-encoder dropout: 0.2202
        \item Classification layer dropout: 0.2457
        \item Learning rate: 0.0004
    \end{itemize}
    \item Amazon HANrnn, Amazon HANconv
    \begin{itemize}
        \item Pre-sentence-encoder dropout: 0.6
        \item Pre-document-encoder dropout: 0.2
        \item Classification layer dropout: 0.4
        \item Learning rate: 0.0002
    \end{itemize}
    \item Amazon HANnoenc
    \begin{itemize}
        \item Pre-sentence-encoder dropout: 0.6
        \item Pre-document-encoder dropout: 0.2
        \item Classification layer dropout: 0.4
        \item Learning rate: 0.0002
        \item Patience: 10
    \end{itemize}
    \item Yelp HANrnn, Yelp HANconv
    \begin{itemize}
        \item Pre-sentence-encoder dropout: 0.7
        \item Pre-document-encoder dropout: 0.1
        \item Classification layer dropout: 0.7
        \item Learning rate: 0.0001
    \end{itemize}
    \item Yelp HANnoenc
    \begin{itemize}
        \item Pre-sentence-encoder dropout: 0.7
        \item Pre-document-encoder dropout: 0.1
        \item Classification layer dropout: 0.7
        \item Learning rate: 0.0001
        \item Patience: 10
    \end{itemize}
    \item Yahoo Answer HANnoenc
    \begin{itemize}
        \item Pre-sentence-encoder dropout: 0.4445
        \item Pre-document-encoder dropout: 0.2202
        \item Classification layer dropout: 0.3749
        \item Learning rate: 0.0004
        \item Patience: 10
    \end{itemize}
    \item IMDB HANconv
    \begin{itemize}
        \item Pre-sentence-encoder dropout: 0.4445
        \item Pre-document-encoder dropout: 0.2202
        \item Classification layer dropout: 0.2457
        \item Learning rate: 0.0004
    \end{itemize}
    \item IMDB HANnoenc
    \begin{itemize}
        \item Pre-sentence-encoder dropout: 0.4445
        \item Pre-document-encoder dropout: 0.2202
        \item Classification layer dropout: 0.2457
        \item Learning rate: 0.0004
        \item Patience: 10
    \end{itemize}
\end{itemize}
For the FLAN models, these are the values we used:
\begin{itemize}
    \item Yahoo Answers FLANrnn, Yahoo Answers FLANconv
    \begin{itemize}
        \item Pre-document-encoder dropout: 0.4445
        \item Classification layer dropout: 0.4457
        \item Learning rate: 0.0004
    \end{itemize}
    \item IMDB FLANrnn, IMDB FLANconv
    \begin{itemize}
        \item Pre-document-encoder dropout: 0.4445
        \item Classification layer dropout: 0.3457
        \item Learning rate: 0.0004
    \end{itemize}
    \item Amazon FLANrnn, Amazon FLANconv
    \begin{itemize}
        \item Pre-document-encoder dropout: 0.6
        \item Classification layer dropout: 0.4
        \item Learning rate: 0.0002
    \end{itemize}
    \item Amazon FLANnoenc
    \begin{itemize}
        \item Pre-document-encoder dropout: 0.6
        \item Classification layer dropout: 0.4
        \item Learning rate: 0.0002
        \item Patience: 10
    \end{itemize}
    \item Yelp FLANrnn, Yelp FLANconv
    \begin{itemize}
        \item Pre-document-encoder dropout: 0.7
        \item Classification layer dropout: 0.7
        \item Learning rate: 0.0001
    \end{itemize}
    \item Yelp FLANnoenc
    \begin{itemize}
        \item Pre-document-encoder dropout: 0.7
        \item Classification layer dropout: 0.7
        \item Learning rate: 0.0001
        \item Patience: 10
    \end{itemize}
    \item Yahoo Answers FLANnoenc
    \begin{itemize}
        \item Pre-document-encoder dropout: 0.4445
        \item Classification layer dropout: 0.4457
        \item Learning rate: 0.0004
        \item Patience: 10
    \end{itemize}
    \item IMDB FLANnoenc
    \begin{itemize}
        \item Pre-document-encoder dropout: 0.4445
        \item Classification layer dropout: 0.3457
        \item Learning rate: 0.0004
        \item Patience: 10
    \end{itemize}
\end{itemize}

Trained model classification accuracies are reported in Table~\ref{model-accuracy}. We note that our IMDB data and Yelp data are different sets of reviews from those used by \citet{han}, so our reported performances are not directly comparable to theirs. We were unable to reach a comparable performance for the Amazon dataset (and Yelp dataset, although different) to that in \cite{han}. We suspect that this is due to not pretraining the word2vec embeddings used by the model for long enough, combined with memory limitations on our hardware that necessitated decreasing our batch size in many cases. However, as noted in section 3, the analysis that we perform does not depend on model accuracy. It's also worth noting that for the datasets for which we \textit{are} able to get results that either pass or come close to the accuracies listed in the original HAN paper, the patterns we see in the results for the tests that we run are the same as the patterns that we see for the others.

\subsection{Full Sets of Plots}

Here we include the full sets of result plots for all models for all tests we describe in the paper, in order of appearance.

\begin{figure*}

\captionsetup[subfloat]{labelformat=empty}

\subfloat{
  \includegraphics[width=\columnwidth]{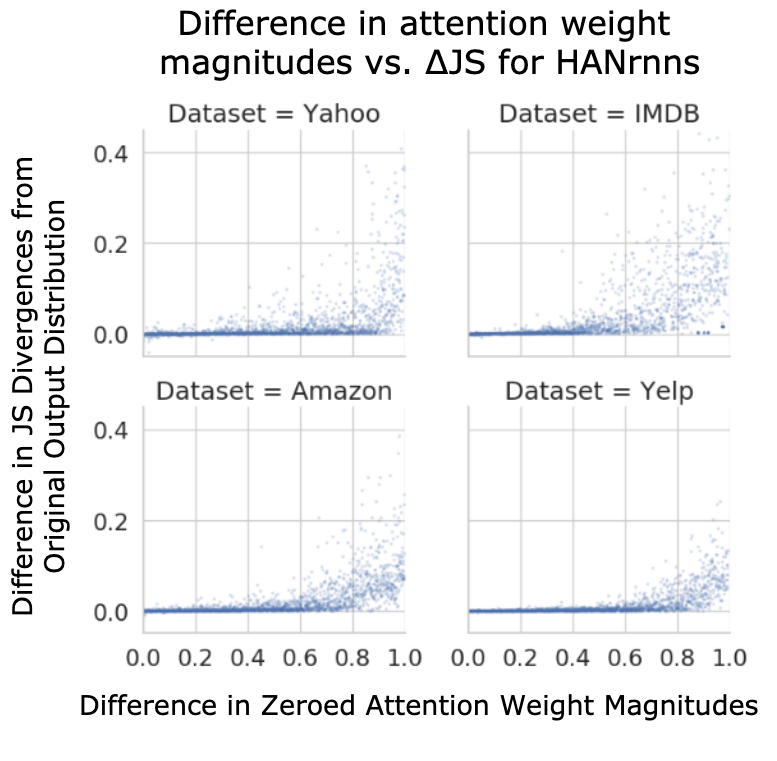}
  \label{app-hanrnn-jsdivscatter}
}%
\subfloat{
  \includegraphics[width=\columnwidth]{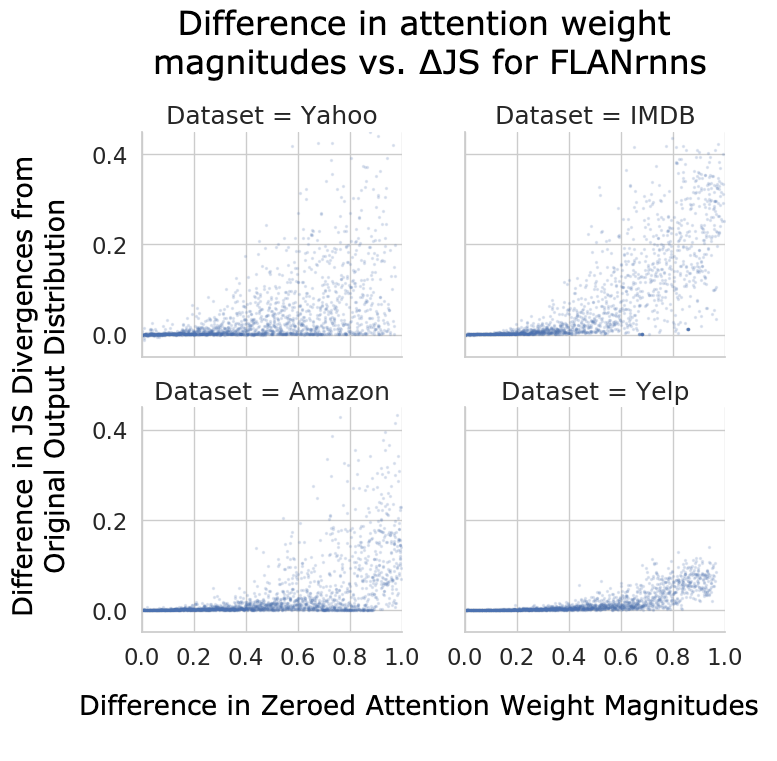}
  \label{app-flanrnn-jsdivscatter}
}\\%
\subfloat{
  \includegraphics[width=\columnwidth]{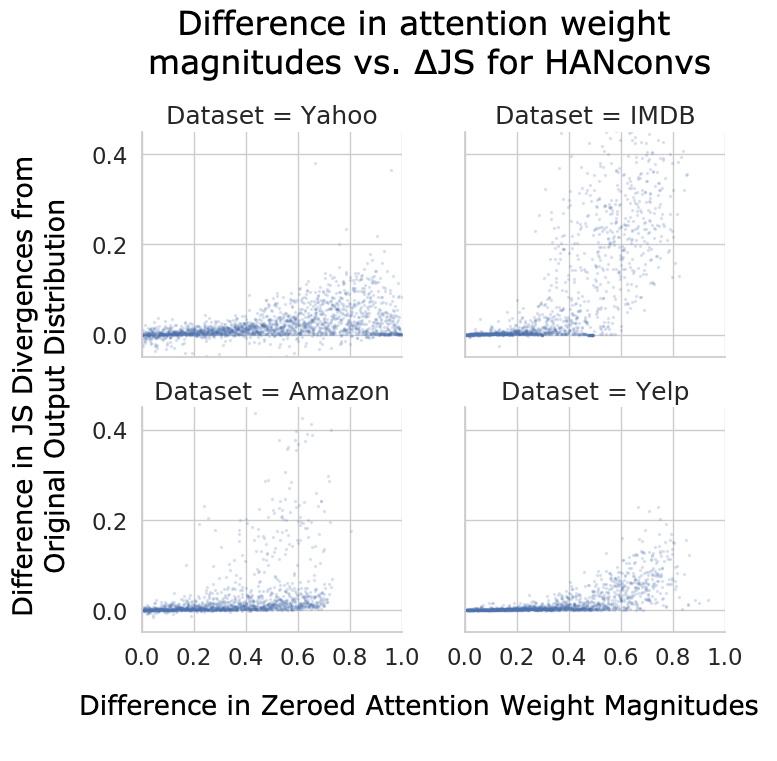}
  \label{app-hanconv-jsdivscatter}
}%
\subfloat{
  \includegraphics[width=\columnwidth]{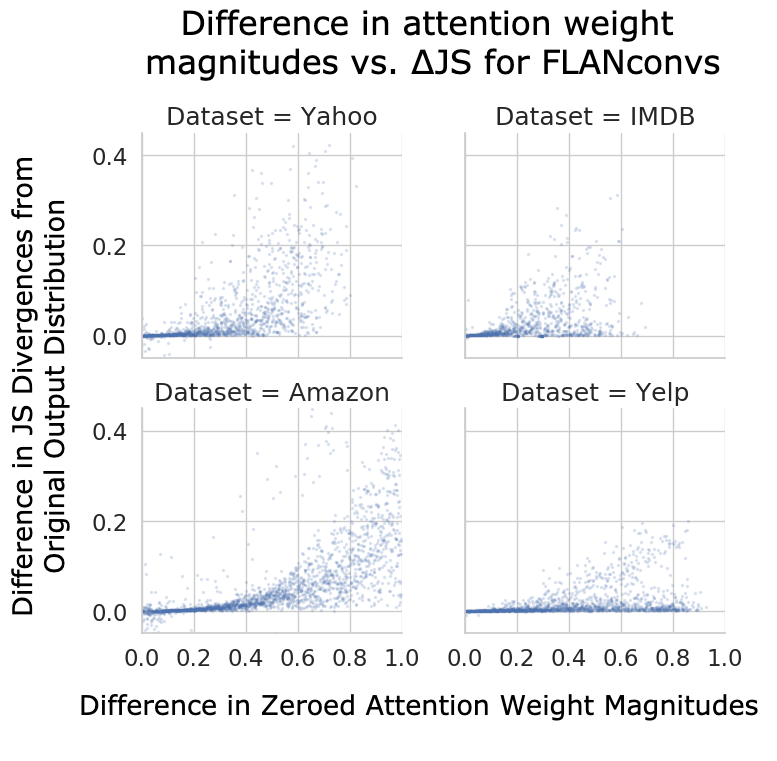}
  \label{app-flanconv-jsdivscatter}
}\\%
\subfloat{
  \includegraphics[width=\columnwidth]{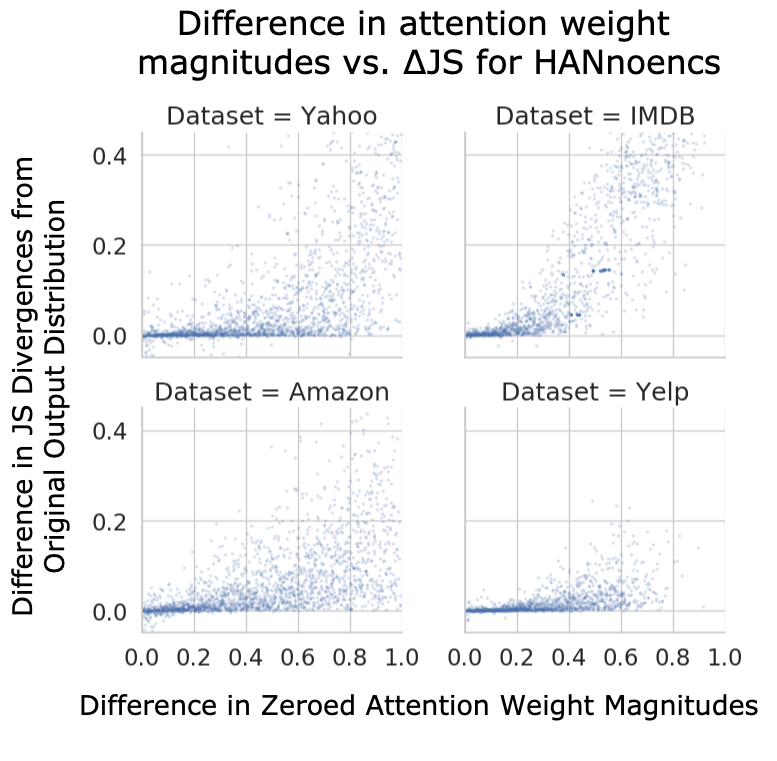}
  \label{app-hannoenc-jsdivscatter}
}%
\subfloat{
  \includegraphics[width=\columnwidth]{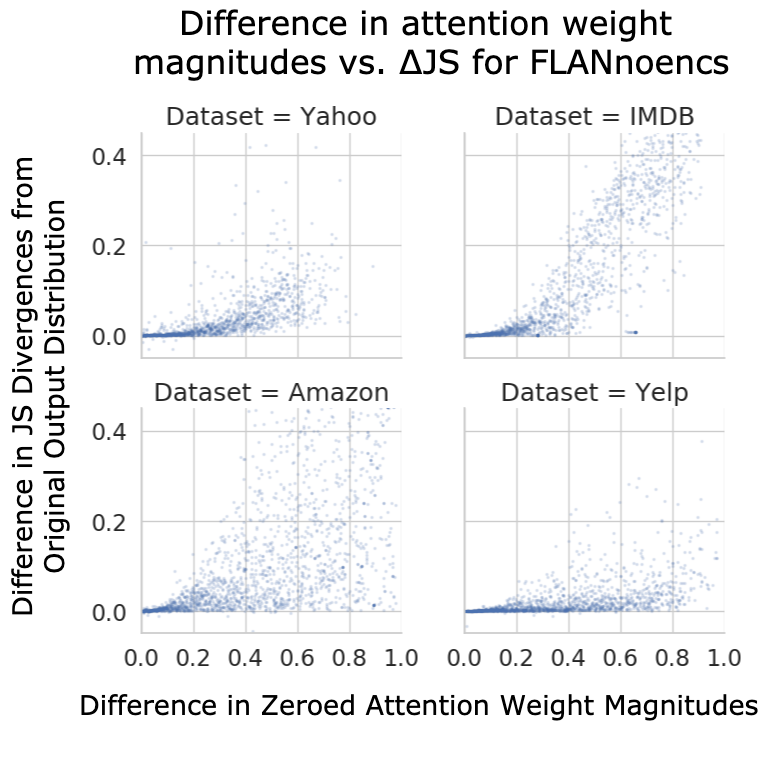}
  \label{app-flannoenc-jsdivscatter}
}%
\caption{Differences in attention weight magnitude plotted against $\Delta$JS for all datasets and architectures}
\label{all-delta-js}
\end{figure*}

In Figure \ref{all-delta-js}, we see that the majority of $\Delta$JS values continue to fall above 0, and that most are still close to 0. One point not stated in the paper, though, is that the upswing in $\Delta $JS values as the difference between $i^\ast$'s weight and a randomly chosen weight increases tends to occur slightly earlier for models with less contextualization, implying that the improving efficiency of the attention-based ranking at flipping the decision as contextualization scope shrinks is also reflected in single-weight test results. 

\begin{figure*}

\subfloat{
  \includegraphics[width=\columnwidth]{appendix_stuff/neg_jsdivdiff_xvals_hanrnn.png}
  \label{app-hanrnn-jsdivhist}
}%
\subfloat{
  \includegraphics[width=\columnwidth]{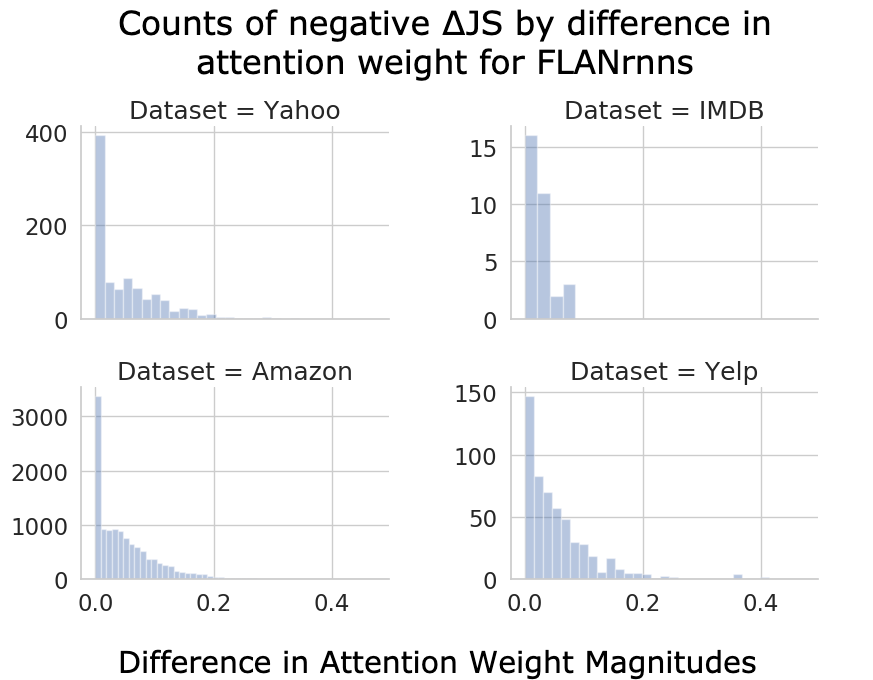}
  \label{app-flanrnn-jsdivhist}
} \\%
\subfloat{
  \includegraphics[width=\columnwidth]{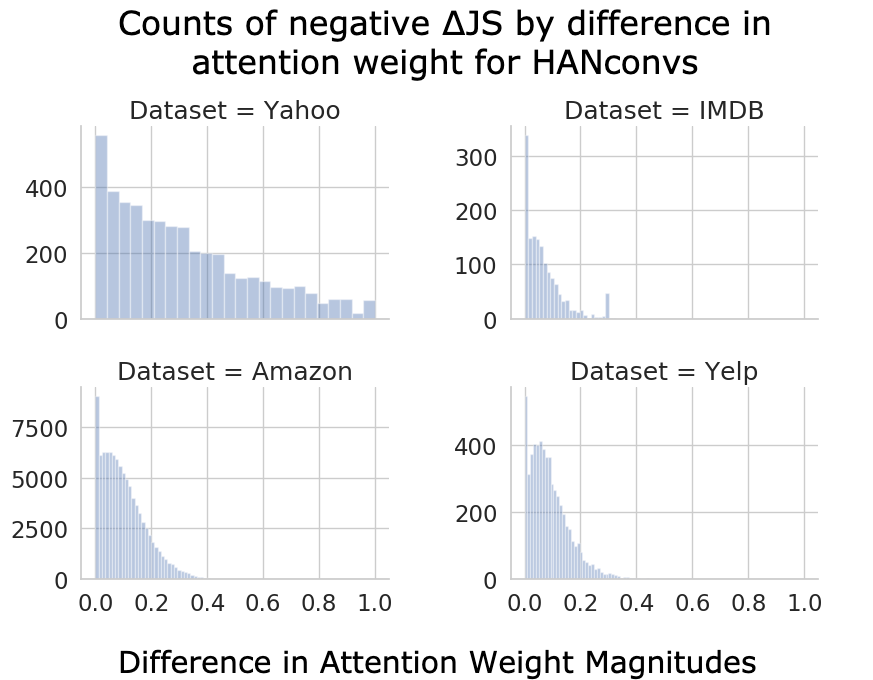}
  \label{app-hanconv-jsdivhist}
}%
\subfloat{
  \includegraphics[width=\columnwidth]{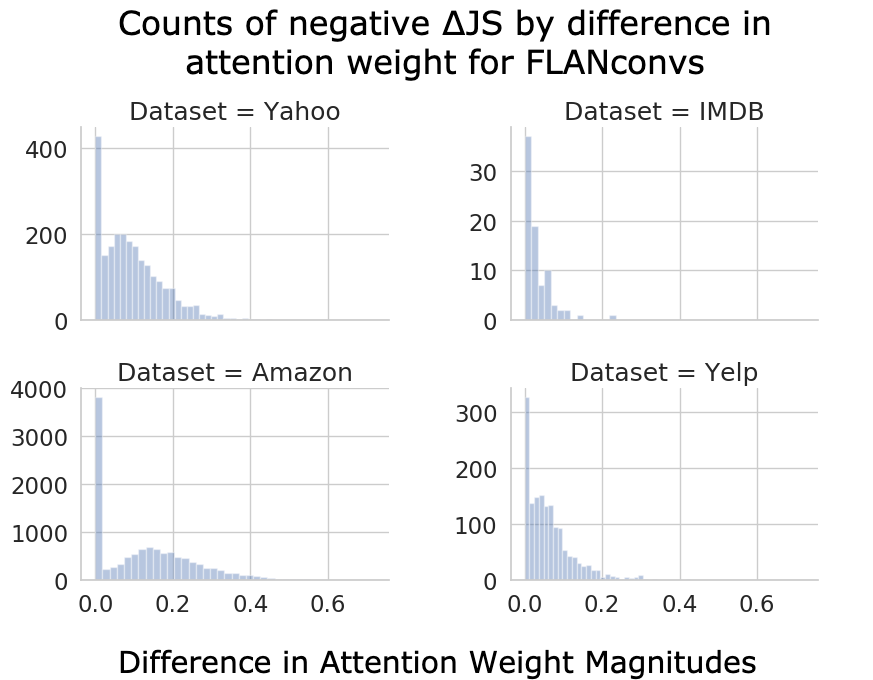}
  \label{app-flanconv-jsdivhist}
}\\%
\subfloat{
  \includegraphics[width=\columnwidth]{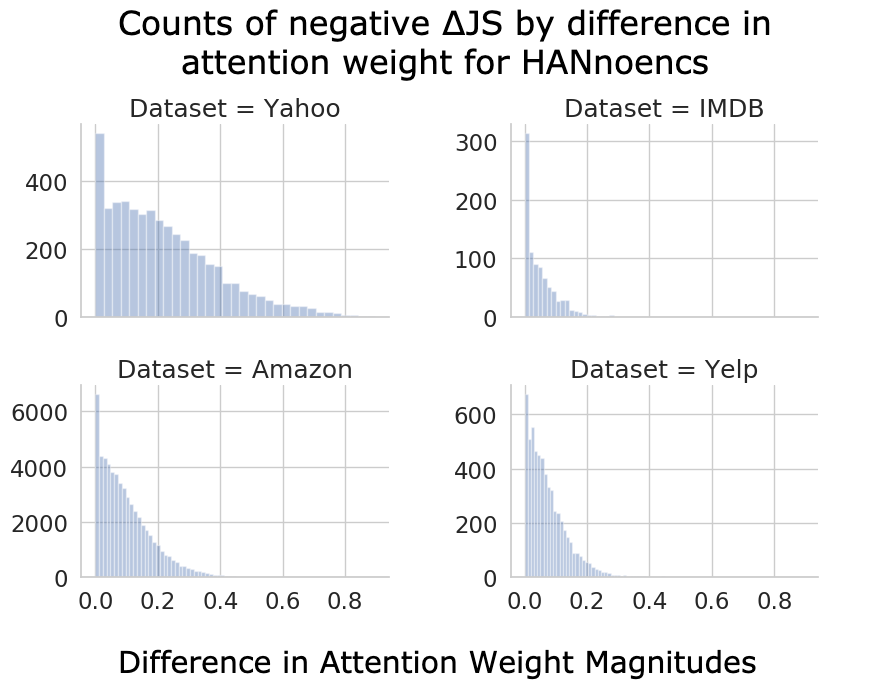}
  \label{app-hannoenc-jsdivhist}
}%
\subfloat{
  \includegraphics[width=\columnwidth]{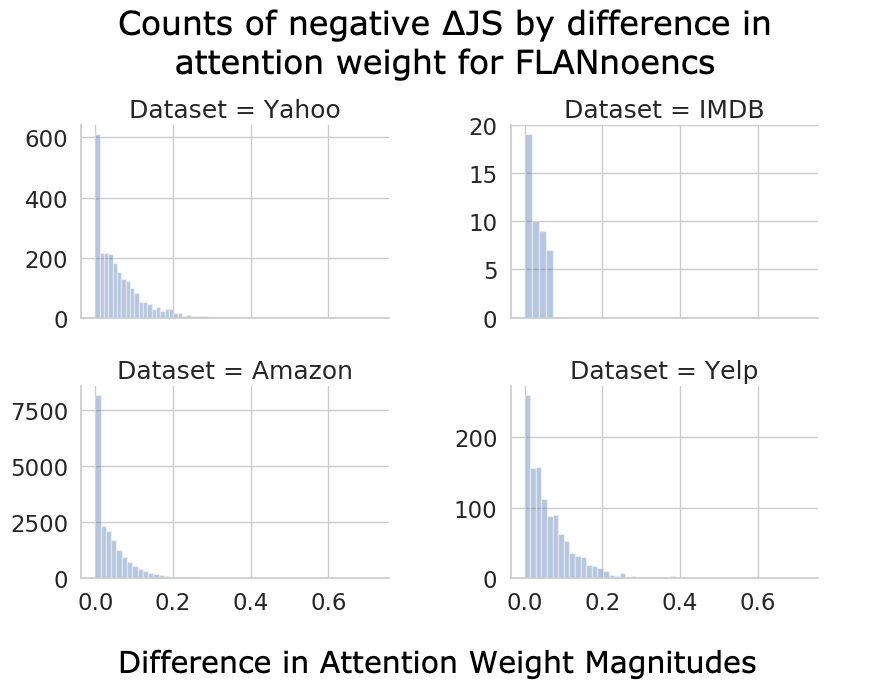}
  \label{app-flannoenc-jsdivhist}
}%
\caption{Counts of negatives $\Delta \mathrm{JS}$ values grouped by the difference in their corresponding attention weights for all datasets and architectures.}
\label{all-delta-js-hists}
\end{figure*}

\begin{figure*}
  \includegraphics[width=\textwidth]{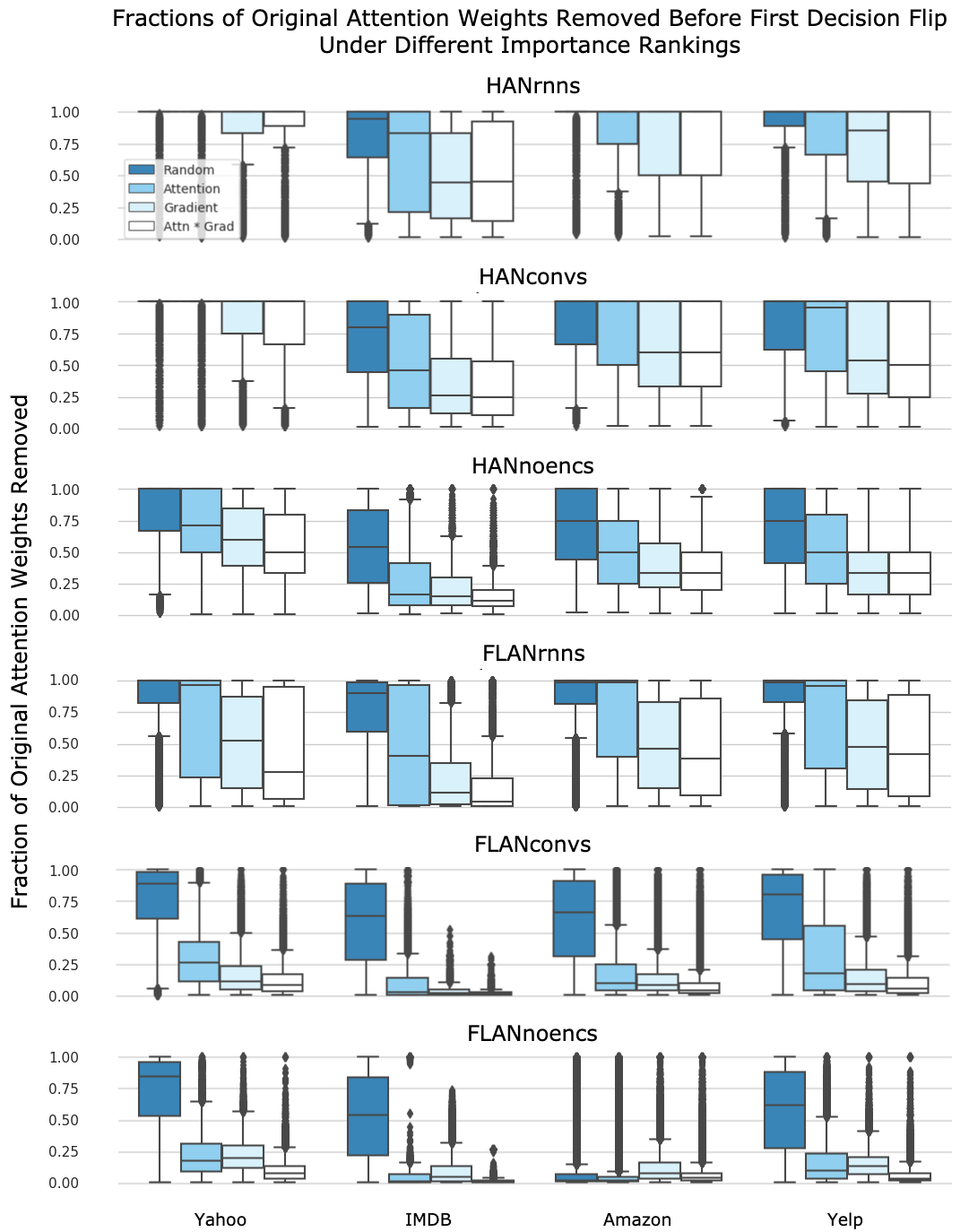}
  \caption{Distribution of fraction of attention weights that had to be removed by different ranking schemes to change each model architecture's decisions for each of the four datasets. The different rankings (aside from ``Attention'', which corresponds to the attention weight magnitudes in descending order) are described in section 5.2.}
  \label{app-fracremoved}
\end{figure*}

Looking at where negative $\Delta$JS values tend to occur in Figure \ref{all-delta-js-hists}, we once again see that they tend to cluster around cases where the difference between the highest and randomly chosen attention weights is close to 0. There are some exceptions, however; perhaps the most obvious are the fat tails of these counts for the Yahoo Answer HAN models. Considering the highest-attention-weight ranking of importance for all Yahoo Answers HAN models in Figure \ref{app-fracremoved} struggle in flipping the decision quickly, it may be that attention is less helpful than usual in identifying importance in its case, which could explain this discrepancy.

\begin{figure*}
    \input{appendix_stuff/decflip_rand_attn_sub_hanrnn.tex}
    \input{appendix_stuff/decflip_rand_attn_sub_flanrnn.tex}\\
    \input{appendix_stuff/decflip_rand_attn_sub_hanconv.tex}
    \input{appendix_stuff/decflip_rand_attn_sub_flanconv.tex}\\
    \input{appendix_stuff/decflip_rand_attn_sub_hannoenc.tex}
    \input{appendix_stuff/decflip_rand_attn_sub_flannoenc.tex}
    \caption{Using the definition of $i^\ast$ given in section 4 (the highest-attention-weight attended item) and comparing to a different randomly selected attended item, these were the percentages of test instances that fell into each decision-flip indicator variable category for each of the four test sets on all models. Since we require our random item not to be $i^\ast$, we exclude any instances with a final sequence length of 1 (one sentence for the HANs, one word for the FLANs) from analysis.}
     \label{dec-flip-tables-attn}
\end{figure*}
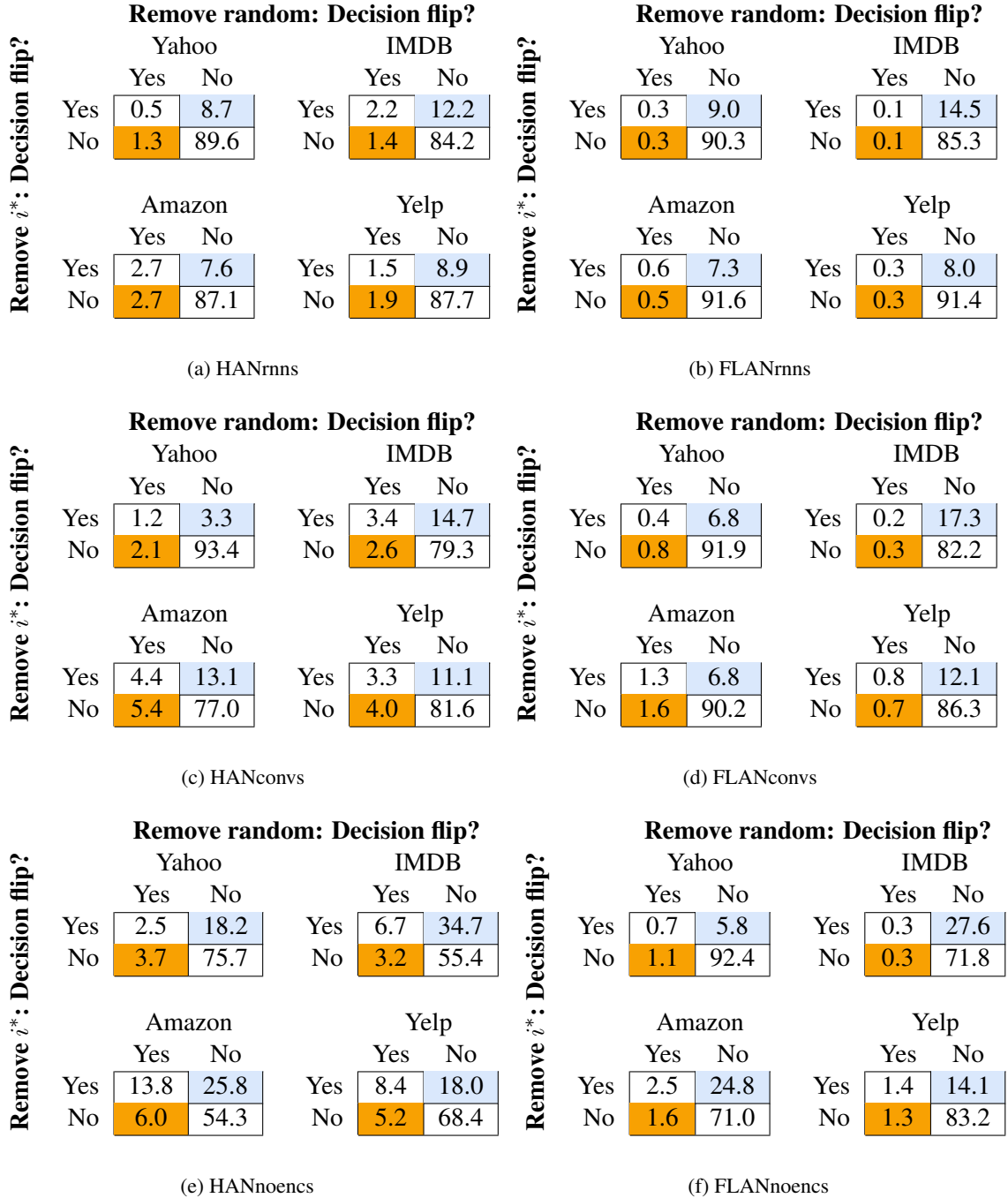

In Figure \ref{dec-flip-tables-attn}, we list contingency tables for all $i^\ast$-versus-random single-weight decision-flip tests. We continue to see higher values overall in our blue cells than orange, as described in section 4.2. The most general change we notice across all the tables is that in the encoderless case, there are more test instances (often many more) where at least one of $i^\ast$ or our random attended item flipped the decision than for any other architecture, except in the case of the Yahoo Answers FLAN. Thinking about why this might be, we recall that in the encoderless case, word embeddings are much more directly responsible for encoding a decision. Yahoo Answers is our only topic classification dataset, where keywords like ``computer'' or ``basketball'' might be much clearer indicators of a topic than, say, ``like'' or ``love'' would be indicators of a rating of 8 versus 9. This likely leads to much less certain decisions being encoded in the word embeddings of the non-Yahoo Answers datasets. For all other models, and in the case where potentially contradictory Yahoo Answers word embeddings are blended together before the final layer of attention (its HANnoenc), it is likely that decisions are simply more brittle overall.

Finally, in Figure \ref{app-fracremoved}, we include the full set of fraction-removed distributions for the first decision flips reached under the different rankings we explored.

\captionsetup[subfloat]{labelformat=parens}
\captionsetup[subfloat]{labelformat=empty}

\captionsetup[subfloat]{labelformat=parens}

\subsection{Additional Tests}

Besides the tests we describe in the main paper, some of the other tests that we ran provide additional insights into our results. We briefly describe those here.

\begin{figure*}
  \includegraphics[width=\textwidth]{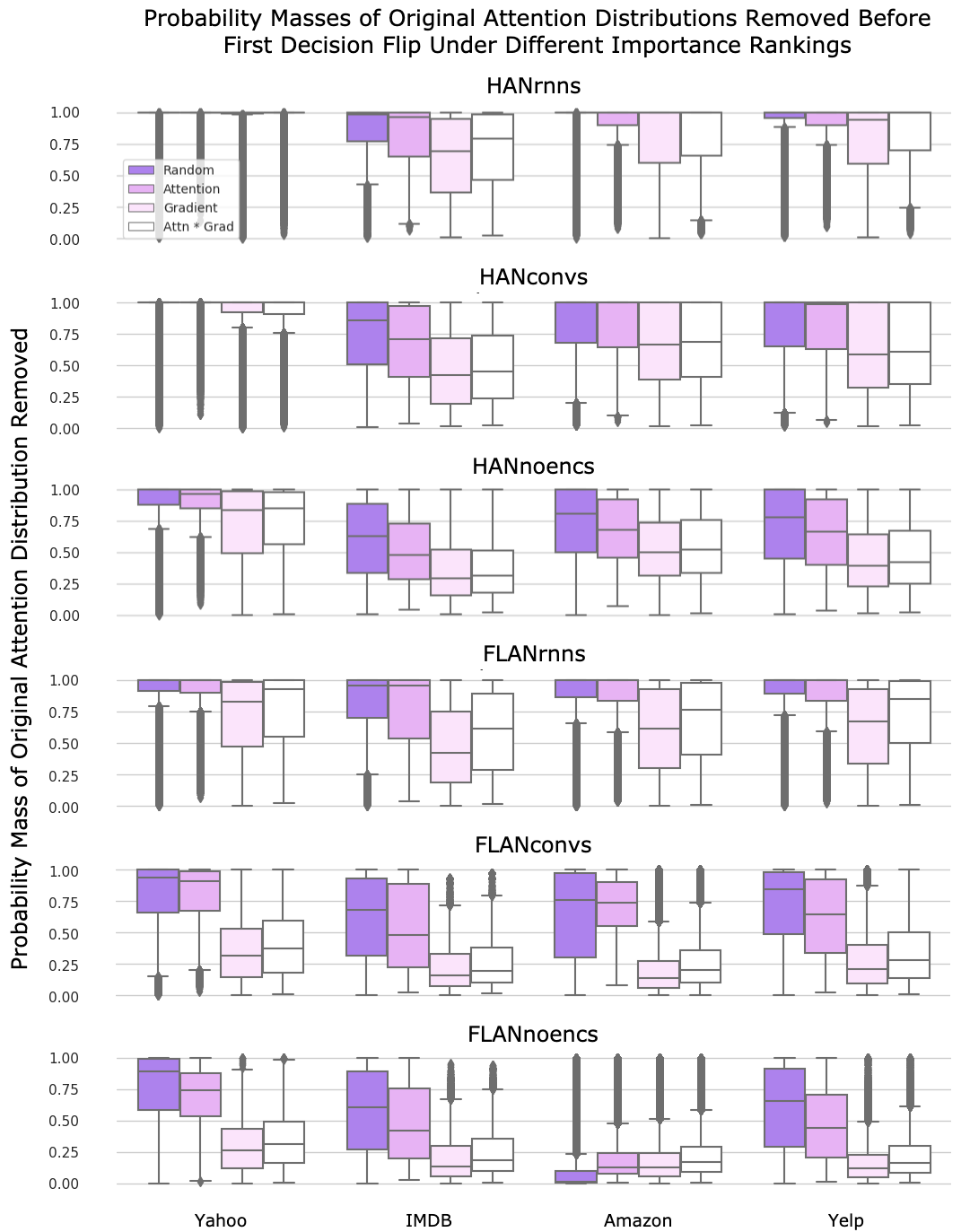}
  \caption{Distribution of probability masses that had to be removed by different ranking schemes to change each model architecture's decisions for each of the four datasets. While we do not discuss these in the paper due to space constraints, we notice that in most cases, a high fraction of the original attention distribution's probability mass must be zeroed before the (renormalized) modified attended representation results in a changed decision using the Attention ranking.}
  \label{app-probmassremoved}
\end{figure*}

In Figure \ref{app-probmassremoved}, we provide the distributions of the original attention \textit{probability distributions} that were zeroed at the point when different ranking schemes achieved their first decision flips. (Equivalently, these are the distributions of the sums of the zeroed attention weights described in Figure \ref{app-fracremoved}, only without repeated normalization.) We include these results to give a sense of \textit{which} attention magnitudes the different rankings typically place towards the top. We notice that this probability mass required to change a decision is often quite high, which is unsurprising for the attention-based ranking, given that it frequently requires removing many items to flip decisions and attention distributions tend to have just a few high weights.

Besides that, the main takeaway that we see here is that for most models, the distribution of attention probability masses zeroed by our gradient-based ranking or our product-based ranking is often shifted down by around 0.25 or more compared to the corresponding attention probability mass distribution for the attention-based ranking, which is a fairly large difference. This would seem to imply that these alternative rankings (which usually flip decisions faster) tend to differ in relatively substantial ways from the rankings suggested by the pure attention weights, not just in the long tail of their orderings, which is another warning sign against attention's interpretability.

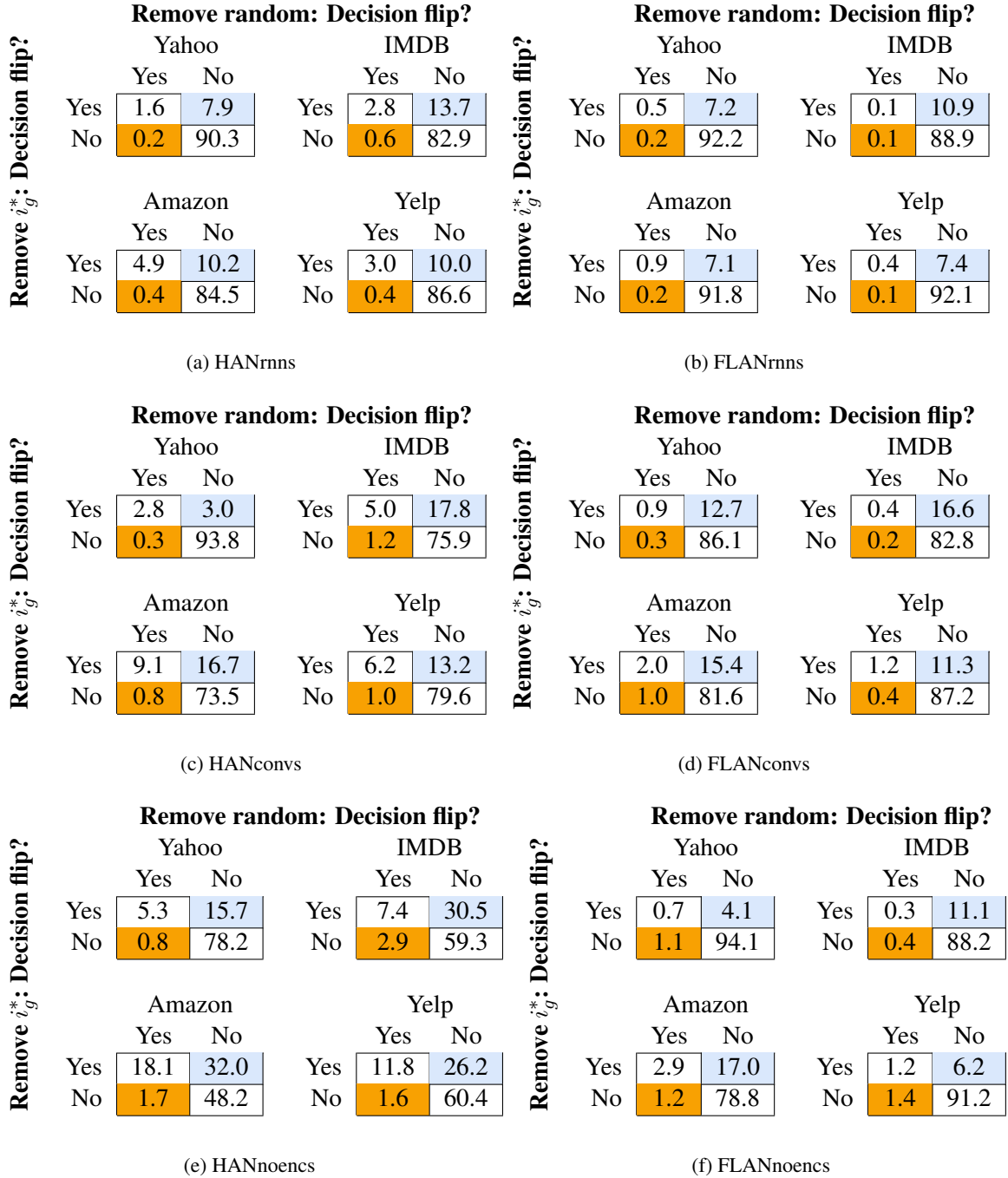
\begin{figure*}
    \input{appendix_stuff/decflip_rand_grad_sub_hanrnn.tex}
    \input{appendix_stuff/decflip_rand_grad_sub_flanrnn.tex}\\
    \input{appendix_stuff/decflip_rand_grad_sub_hanconv.tex}
    \input{appendix_stuff/decflip_rand_grad_sub_flanconv.tex}\\
    \input{appendix_stuff/decflip_rand_grad_sub_hannoenc.tex}
    \input{appendix_stuff/decflip_rand_grad_sub_flannoenc.tex}
    \caption{Let $i^\ast_g$ be the highest-ranked attended item using our purely gradient-based ranking of importance described in section 5.2. We rerun our single-weight decision flip tests using this new $i^\ast_g$, comparing to a different randomly selected attended item. These were the percentages of test instances that fell into each decision-flip indicator variable category for each of the four test sets on all models. Since we require our random item not to be $i^\ast_g$, we exclude any instances with a final sequence length of 1 (one sentence for the HANs, one word for the FLANs) from analysis.}
     \label{dec-flip-tables-grad}
\end{figure*}

\begin{figure*}
    \input{appendix_stuff/decflip_rand_gradmult_sub_hanrnn.tex}
    \input{appendix_stuff/decflip_rand_gradmult_sub_flanrnn.tex}\\
    \input{appendix_stuff/decflip_rand_gradmult_sub_hanconv.tex}
    \input{appendix_stuff/decflip_rand_gradmult_sub_flanconv.tex}\\
    \input{appendix_stuff/decflip_rand_gradmult_sub_hannoenc.tex}
    \input{appendix_stuff/decflip_rand_gradmult_sub_flannoenc.tex}
    \caption{Let $i^\ast_p$ be the highest-ranked attended item using our attention-gradient product ranking of importance described in section 5.2. Once again, we rerun our single-weight decision flip tests using this new $i^\ast_p$, comparing to a different randomly selected attended item. These were the percentages of test instances that fell into each decision-flip indicator variable category for each of the four test sets on all models. Since we require our random item not to be $i^\ast_p$, we exclude any instances with a final sequence length of 1 (one sentence for the HANs, one word for the FLANs) from analysis.}
     \label{dec-flip-tables-gradmult}
\end{figure*}
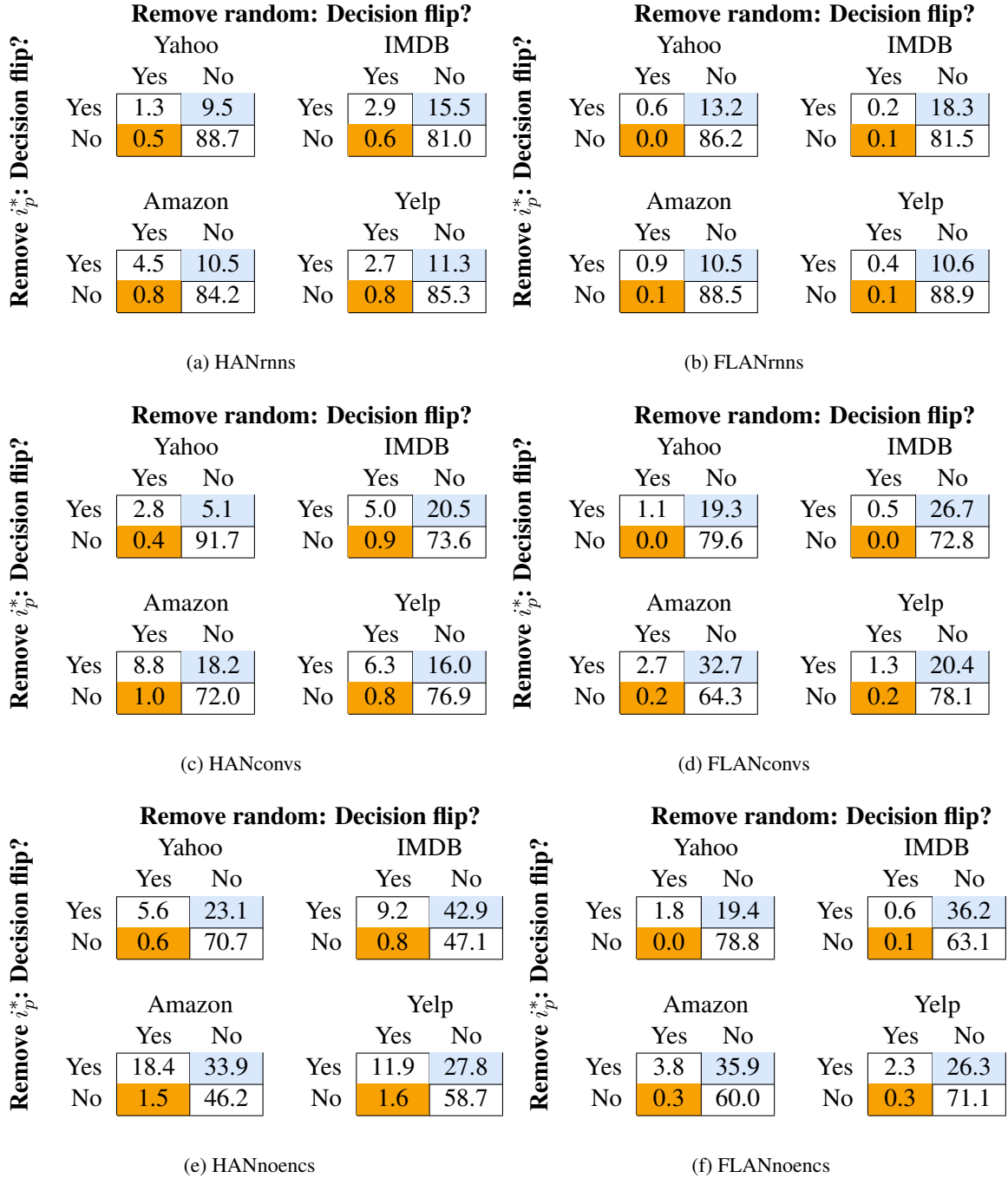

The final set of tests that we include in Figures \ref{dec-flip-tables-grad} and \ref{dec-flip-tables-gradmult} consist of rerunning our single-weight decision-flip tests on the single ``most important'' attention weights in their respective attention distributions as suggested by our \textit{alternative} rankings (gradient-based and product-based rankings) instead of attention magnitudes. These results serve two functions: first, they imply still more information about when the top weight suggested by an alternative, faster-decision-flipping ranking differs from the top attention weight. Intuitively, if we observe large differences between the sum of the ``yes'' row for one contingency table and the ``yes'' rows for the other rankings' tables on that same model, this is likely due to differences in the frequencies with which the highest-ranked items achieve a decision flip, indicating differences in highest-ranked items (``likely'' because of the noise added by the random sampling).

The second piece of information that these tests provide is a lower bound (via the sum of the ``yes'' rows) for the number of cases where rankings flip a decision as quickly as possible (i.e., in the first item). For context, the sum of the ``yes'' row is higher than the corresponding sum in Figure \ref{dec-flip-tables-attn} for \textit{all} contingency tables using our product-based ordering. For the gradient-based ordering, however, this sum is actually lower than for the attention-based ranking's tables in 14 out of our 24 models. This tells us that our gradient-based method often finds fewer single-item ways of flipping decisions than the attention-based ranking, so in order to achieve its more efficient overall distribution of flips that we see for many models in Figure \ref{app-fracremoved}, it must usually flip decisions faster than attention in cases where both its ranking and the attention-based ranking require multiple removed weights.

%% file: appendix_stuff/model_performance_table.tex
\begin{table*}[t!]
\begin{center}
\begin{tabular}{lcccccc}
\hline \bf Dataset & \bf HANrnn & \bf HANconv & \bf HANnoenc & \bf FLANrnn & \bf FLANconv & \bf FLANnoenc \\ \hline
Yahoo Answers & 74.6 & 72.8 & 73.1 & 75.5 & 73.1 & 72.3 \\
IMDB & 50.3 & 48.9 & 46.1 & 49.1 & 48.2 & 45.4 \\
Amazon & 56.9 & 55.3 & 51.2 & 56.6 & 54.4 & 50.2 \\
Yelp & 63.0 & 61.0 & 58.6 & 62.3 & 60.7 & 58.2 \\
\hline
\end{tabular}
\end{center}
\caption{\label{model-accuracy} Classification accuracy of the different trained models on their respective test sets}
\end{table*}

%% file: appendix_stuff/decflip_rand_attn_sub_hanrnn.tex
\subfloat[HANrnns]{
\begin{tabular}{clccllcc}
\multicolumn{1}{l}{}                &      & \multicolumn{6}{c}{\textbf{Remove random: Decision flip?}}                                                                                                                                                                                                                  \\
                                          &                           & \multicolumn{2}{c}{Yahoo}                                                                       &  &                           & \multicolumn{2}{c}{IMDB}                                                                        \\
                                          &                           & Yes                                           & No                                           &  &                           & Yes                                           & No                                           \\ \cline{3-4} \cline{7-8} 
                                          & \multicolumn{1}{r|}{Yes} & \multicolumn{1}{c|}{0.5}                         & \multicolumn{1}{c|}{\cellcolor[HTML]{DAE8FC}8.7} &  & \multicolumn{1}{r|}{Yes} & \multicolumn{1}{c|}{2.2}                         & \multicolumn{1}{c|}{\cellcolor[HTML]{DAE8FC}12.2} \\ \cline{3-4} \cline{7-8} 
                                          & \multicolumn{1}{r|}{No} & \multicolumn{1}{c|}{\cellcolor[HTML]{F8A102}1.3} & \multicolumn{1}{c|}{89.6}                         &  & \multicolumn{1}{r|}{No} & \multicolumn{1}{c|}{\cellcolor[HTML]{F8A102}1.4} & \multicolumn{1}{c|}{84.2}                         \\ \cline{3-4} \cline{7-8} 
                                          &                           & \multicolumn{1}{l}{}                           & \multicolumn{1}{l}{}                           &  &                           & \multicolumn{1}{l}{}                           & \multicolumn{1}{l}{}                           \\
\end{tabular}
}

%% file: appendix_stuff/decflip_rand_attn_sub_flanrnn.tex
\subfloat[FLANrnns]{
\begin{tabular}{clccllcc}
\multicolumn{1}{l}{}                &      & \multicolumn{6}{c}{\textbf{Remove random: Decision flip?}}                                                                                                                                                                                                                  \\
                                          &                           & \multicolumn{2}{c}{Yahoo}                                                                       &  &                           & \multicolumn{2}{c}{IMDB}                                                                        \\
                                          &                           & Yes                                           & No                                           &  &                           & Yes                                           & No                                           \\ \cline{3-4} \cline{7-8} 
                                          & \multicolumn{1}{r|}{Yes} & \multicolumn{1}{c|}{0.3}                         & \multicolumn{1}{c|}{\cellcolor[HTML]{DAE8FC}9.0} &  & \multicolumn{1}{r|}{Yes} & \multicolumn{1}{c|}{0.1}                         & \multicolumn{1}{c|}{\cellcolor[HTML]{DAE8FC}14.5} \\ \cline{3-4} \cline{7-8} 
                                          & \multicolumn{1}{r|}{No} & \multicolumn{1}{c|}{\cellcolor[HTML]{F8A102}0.3} & \multicolumn{1}{c|}{90.3}                         &  & \multicolumn{1}{r|}{No} & \multicolumn{1}{c|}{\cellcolor[HTML]{F8A102}0.1} & \multicolumn{1}{c|}{85.3}                         \\ \cline{3-4} \cline{7-8} 
                                          &                           & \multicolumn{1}{l}{}                           & \multicolumn{1}{l}{}                           &  &                           & \multicolumn{1}{l}{}                           & \multicolumn{1}{l}{}                           \\
\end{tabular}
}

%% file: appendix_stuff/decflip_rand_attn_sub_hanconv.tex
\subfloat[HANconvs]{
\begin{tabular}{clccllcc}
\multicolumn{1}{l}{}                &      & \multicolumn{6}{c}{\textbf{Remove random: Decision flip?}}                                                                                                                                                                                                                  \\
                                          &                           & \multicolumn{2}{c}{Yahoo}                                                                       &  &                           & \multicolumn{2}{c}{IMDB}                                                                        \\
                                          &                           & Yes                                           & No                                           &  &                           & Yes                                           & No                                           \\ \cline{3-4} \cline{7-8} 
                                          & \multicolumn{1}{r|}{Yes} & \multicolumn{1}{c|}{1.2}                         & \multicolumn{1}{c|}{\cellcolor[HTML]{DAE8FC}3.3} &  & \multicolumn{1}{r|}{Yes} & \multicolumn{1}{c|}{3.4}                         & \multicolumn{1}{c|}{\cellcolor[HTML]{DAE8FC}14.7} \\ \cline{3-4} \cline{7-8} 
                                          & \multicolumn{1}{r|}{No} & \multicolumn{1}{c|}{\cellcolor[HTML]{F8A102}2.1} & \multicolumn{1}{c|}{93.4}                         &  & \multicolumn{1}{r|}{No} & \multicolumn{1}{c|}{\cellcolor[HTML]{F8A102}2.6} & \multicolumn{1}{c|}{79.3}                         \\ \cline{3-4} \cline{7-8} 
                                          &                           & \multicolumn{1}{l}{}                           & \multicolumn{1}{l}{}                           &  &                           & \multicolumn{1}{l}{}                           & \multicolumn{1}{l}{}                           \\
\end{tabular}
}

%% file: appendix_stuff/decflip_rand_attn_sub_flanconv.tex
\subfloat[FLANconvs]{
\begin{tabular}{clccllcc}
\multicolumn{1}{l}{}                &      & \multicolumn{6}{c}{\textbf{Remove random: Decision flip?}}                                                                                                                                                                                                                  \\
                                          &                           & \multicolumn{2}{c}{Yahoo}                                                                       &  &                           & \multicolumn{2}{c}{IMDB}                                                                        \\
                                          &                           & Yes                                           & No                                           &  &                           & Yes                                           & No                                           \\ \cline{3-4} \cline{7-8} 
                                          & \multicolumn{1}{r|}{Yes} & \multicolumn{1}{c|}{0.4}                         & \multicolumn{1}{c|}{\cellcolor[HTML]{DAE8FC}6.8} &  & \multicolumn{1}{r|}{Yes} & \multicolumn{1}{c|}{0.2}                         & \multicolumn{1}{c|}{\cellcolor[HTML]{DAE8FC}17.3} \\ \cline{3-4} \cline{7-8} 
                                          & \multicolumn{1}{r|}{No} & \multicolumn{1}{c|}{\cellcolor[HTML]{F8A102}0.8} & \multicolumn{1}{c|}{91.9}                         &  & \multicolumn{1}{r|}{No} & \multicolumn{1}{c|}{\cellcolor[HTML]{F8A102}0.3} & \multicolumn{1}{c|}{82.2}                         \\ \cline{3-4} \cline{7-8} 
                                          &                           & \multicolumn{1}{l}{}                           & \multicolumn{1}{l}{}                           &  &                           & \multicolumn{1}{l}{}                           & \multicolumn{1}{l}{}                           \\
\end{tabular}
}

%% file: appendix_stuff/decflip_rand_attn_sub_hannoenc.tex
\subfloat[HANnoencs]{
\begin{tabular}{clccllcc}
\multicolumn{1}{l}{}                &      & \multicolumn{6}{c}{\textbf{Remove random: Decision flip?}}                                                                                                                                                                                                                  \\
                                          &                           & \multicolumn{2}{c}{Yahoo}                                                                       &  &                           & \multicolumn{2}{c}{IMDB}                                                                        \\
                                          &                           & Yes                                           & No                                           &  &                           & Yes                                           & No                                           \\ \cline{3-4} \cline{7-8} 
                                          & \multicolumn{1}{r|}{Yes} & \multicolumn{1}{c|}{2.5}                         & \multicolumn{1}{c|}{\cellcolor[HTML]{DAE8FC}18.2} &  & \multicolumn{1}{r|}{Yes} & \multicolumn{1}{c|}{6.7}                         & \multicolumn{1}{c|}{\cellcolor[HTML]{DAE8FC}34.7} \\ \cline{3-4} \cline{7-8} 
                                          & \multicolumn{1}{r|}{No} & \multicolumn{1}{c|}{\cellcolor[HTML]{F8A102}3.7} & \multicolumn{1}{c|}{75.7}                         &  & \multicolumn{1}{r|}{No} & \multicolumn{1}{c|}{\cellcolor[HTML]{F8A102}3.2} & \multicolumn{1}{c|}{55.4}                         \\ \cline{3-4} \cline{7-8} 
                                          &                           & \multicolumn{1}{l}{}                           & \multicolumn{1}{l}{}                           &  &                           & \multicolumn{1}{l}{}                           & \multicolumn{1}{l}{}                           \\
\end{tabular}
}

%% file: appendix_stuff/decflip_rand_attn_sub_flannoenc.tex
\subfloat[FLANnoencs]{
\begin{tabular}{clccllcc}
\multicolumn{1}{l}{}                &      & \multicolumn{6}{c}{\textbf{Remove random: Decision flip?}}                                                                                                                                                                                                                  \\
                                          &                           & \multicolumn{2}{c}{Yahoo}                                                                       &  &                           & \multicolumn{2}{c}{IMDB}                                                                        \\
                                          &                           & Yes                                           & No                                           &  &                           & Yes                                           & No                                           \\ \cline{3-4} \cline{7-8} 
                                          & \multicolumn{1}{r|}{Yes} & \multicolumn{1}{c|}{0.7}                         & \multicolumn{1}{c|}{\cellcolor[HTML]{DAE8FC}5.8} &  & \multicolumn{1}{r|}{Yes} & \multicolumn{1}{c|}{0.3}                         & \multicolumn{1}{c|}{\cellcolor[HTML]{DAE8FC}27.6} \\ \cline{3-4} \cline{7-8} 
                                          & \multicolumn{1}{r|}{No} & \multicolumn{1}{c|}{\cellcolor[HTML]{F8A102}1.1} & \multicolumn{1}{c|}{92.4}                         &  & \multicolumn{1}{r|}{No} & \multicolumn{1}{c|}{\cellcolor[HTML]{F8A102}0.3} & \multicolumn{1}{c|}{71.8}                         \\ \cline{3-4} \cline{7-8} 
                                          &                           & \multicolumn{1}{l}{}                           & \multicolumn{1}{l}{}                           &  &                           & \multicolumn{1}{l}{}                           & \multicolumn{1}{l}{}                           \\
\end{tabular}
}

%% file: appendix_stuff/decflip_rand_grad_sub_hanrnn.tex
\subfloat[HANrnns]{
\begin{tabular}{clccllcc}
\multicolumn{1}{l}{}                &      & \multicolumn{6}{c}{\textbf{Remove random: Decision flip?}}                                                                                                                                                                                                                  \\
                                          &                           & \multicolumn{2}{c}{Yahoo}                                                                       &  &                           & \multicolumn{2}{c}{IMDB}                                                                        \\
                                          &                           & Yes                                           & No                                           &  &                           & Yes                                           & No                                           \\ \cline{3-4} \cline{7-8} 
                                          & \multicolumn{1}{r|}{Yes} & \multicolumn{1}{c|}{1.6}                         & \multicolumn{1}{c|}{\cellcolor[HTML]{DAE8FC}7.9} &  & \multicolumn{1}{r|}{Yes} & \multicolumn{1}{c|}{2.8}                         & \multicolumn{1}{c|}{\cellcolor[HTML]{DAE8FC}13.7} \\ \cline{3-4} \cline{7-8} 
                                          & \multicolumn{1}{r|}{No} & \multicolumn{1}{c|}{\cellcolor[HTML]{F8A102}0.2} & \multicolumn{1}{c|}{90.3}                         &  & \multicolumn{1}{r|}{No} & \multicolumn{1}{c|}{\cellcolor[HTML]{F8A102}0.6} & \multicolumn{1}{c|}{82.9}                         \\ \cline{3-4} \cline{7-8} 
                                          &                           & \multicolumn{1}{l}{}                           & \multicolumn{1}{l}{}                           &  &                           & \multicolumn{1}{l}{}                           & \multicolumn{1}{l}{}                           \\
\end{tabular}
}

%% file: appendix_stuff/decflip_rand_grad_sub_flanrnn.tex
\subfloat[FLANrnns]{
\begin{tabular}{clccllcc}
\multicolumn{1}{l}{}                &      & \multicolumn{6}{c}{\textbf{Remove random: Decision flip?}}                                                                                                                                                                                                                  \\
                                          &                           & \multicolumn{2}{c}{Yahoo}                                                                       &  &                           & \multicolumn{2}{c}{IMDB}                                                                        \\
                                          &                           & Yes                                           & No                                           &  &                           & Yes                                           & No                                           \\ \cline{3-4} \cline{7-8} 
                                          & \multicolumn{1}{r|}{Yes} & \multicolumn{1}{c|}{0.5}                         & \multicolumn{1}{c|}{\cellcolor[HTML]{DAE8FC}7.2} &  & \multicolumn{1}{r|}{Yes} & \multicolumn{1}{c|}{0.1}                         & \multicolumn{1}{c|}{\cellcolor[HTML]{DAE8FC}10.9} \\ \cline{3-4} \cline{7-8} 
                                          & \multicolumn{1}{r|}{No} & \multicolumn{1}{c|}{\cellcolor[HTML]{F8A102}0.2} & \multicolumn{1}{c|}{92.2}                         &  & \multicolumn{1}{r|}{No} & \multicolumn{1}{c|}{\cellcolor[HTML]{F8A102}0.1} & \multicolumn{1}{c|}{88.9}                         \\ \cline{3-4} \cline{7-8} 
                                          &                           & \multicolumn{1}{l}{}                           & \multicolumn{1}{l}{}                           &  &                           & \multicolumn{1}{l}{}                           & \multicolumn{1}{l}{}                           \\
\end{tabular}
}

%% file: appendix_stuff/decflip_rand_grad_sub_hanconv.tex
\subfloat[HANconvs]{
\begin{tabular}{clccllcc}
\multicolumn{1}{l}{}                &      & \multicolumn{6}{c}{\textbf{Remove random: Decision flip?}}                                                                                                                                                                                                                  \\
                                          &                           & \multicolumn{2}{c}{Yahoo}                                                                       &  &                           & \multicolumn{2}{c}{IMDB}                                                                        \\
                                          &                           & Yes                                           & No                                           &  &                           & Yes                                           & No                                           \\ \cline{3-4} \cline{7-8} 
                                          & \multicolumn{1}{r|}{Yes} & \multicolumn{1}{c|}{2.8}                         & \multicolumn{1}{c|}{\cellcolor[HTML]{DAE8FC}3.0} &  & \multicolumn{1}{r|}{Yes} & \multicolumn{1}{c|}{5.0}                         & \multicolumn{1}{c|}{\cellcolor[HTML]{DAE8FC}17.8} \\ \cline{3-4} \cline{7-8} 
                                          & \multicolumn{1}{r|}{No} & \multicolumn{1}{c|}{\cellcolor[HTML]{F8A102}0.3} & \multicolumn{1}{c|}{93.8}                         &  & \multicolumn{1}{r|}{No} & \multicolumn{1}{c|}{\cellcolor[HTML]{F8A102}1.2} & \multicolumn{1}{c|}{75.9}                         \\ \cline{3-4} \cline{7-8} 
                                          &                           & \multicolumn{1}{l}{}                           & \multicolumn{1}{l}{}                           &  &                           & \multicolumn{1}{l}{}                           & \multicolumn{1}{l}{}                           \\
\end{tabular}
}

%% file: appendix_stuff/decflip_rand_grad_sub_flanconv.tex
\subfloat[FLANconvs]{
\begin{tabular}{clccllcc}
\multicolumn{1}{l}{}                &      & \multicolumn{6}{c}{\textbf{Remove random: Decision flip?}}                                                                                                                                                                                                                  \\
                                          &                           & \multicolumn{2}{c}{Yahoo}                                                                       &  &                           & \multicolumn{2}{c}{IMDB}                                                                        \\
                                          &                           & Yes                                           & No                                           &  &                           & Yes                                           & No                                           \\ \cline{3-4} \cline{7-8} 
                                          & \multicolumn{1}{r|}{Yes} & \multicolumn{1}{c|}{0.9}                         & \multicolumn{1}{c|}{\cellcolor[HTML]{DAE8FC}12.7} &  & \multicolumn{1}{r|}{Yes} & \multicolumn{1}{c|}{0.4}                         & \multicolumn{1}{c|}{\cellcolor[HTML]{DAE8FC}16.6} \\ \cline{3-4} \cline{7-8} 
                                          & \multicolumn{1}{r|}{No} & \multicolumn{1}{c|}{\cellcolor[HTML]{F8A102}0.3} & \multicolumn{1}{c|}{86.1}                         &  & \multicolumn{1}{r|}{No} & \multicolumn{1}{c|}{\cellcolor[HTML]{F8A102}0.2} & \multicolumn{1}{c|}{82.8}                         \\ \cline{3-4} \cline{7-8} 
                                          &                           & \multicolumn{1}{l}{}                           & \multicolumn{1}{l}{}                           &  &                           & \multicolumn{1}{l}{}                           & \multicolumn{1}{l}{}                           \\
\end{tabular}
}

%% file: appendix_stuff/decflip_rand_grad_sub_hannoenc.tex
\subfloat[HANnoencs]{
\begin{tabular}{clccllcc}
\multicolumn{1}{l}{}                &      & \multicolumn{6}{c}{\textbf{Remove random: Decision flip?}}                                                                                                                                                                                                                  \\
                                          &                           & \multicolumn{2}{c}{Yahoo}                                                                       &  &                           & \multicolumn{2}{c}{IMDB}                                                                        \\
                                          &                           & Yes                                           & No                                           &  &                           & Yes                                           & No                                           \\ \cline{3-4} \cline{7-8} 
                                          & \multicolumn{1}{r|}{Yes} & \multicolumn{1}{c|}{5.3}                         & \multicolumn{1}{c|}{\cellcolor[HTML]{DAE8FC}15.7} &  & \multicolumn{1}{r|}{Yes} & \multicolumn{1}{c|}{7.4}                         & \multicolumn{1}{c|}{\cellcolor[HTML]{DAE8FC}30.5} \\ \cline{3-4} \cline{7-8} 
                                          & \multicolumn{1}{r|}{No} & \multicolumn{1}{c|}{\cellcolor[HTML]{F8A102}0.8} & \multicolumn{1}{c|}{78.2}                         &  & \multicolumn{1}{r|}{No} & \multicolumn{1}{c|}{\cellcolor[HTML]{F8A102}2.9} & \multicolumn{1}{c|}{59.3}                         \\ \cline{3-4} \cline{7-8} 
                                          &                           & \multicolumn{1}{l}{}                           & \multicolumn{1}{l}{}                           &  &                           & \multicolumn{1}{l}{}                           & \multicolumn{1}{l}{}                           \\
\end{tabular}
}

%% file: appendix_stuff/decflip_rand_grad_sub_flannoenc.tex
\subfloat[FLANnoencs]{
\begin{tabular}{clccllcc}
\multicolumn{1}{l}{}                &      & \multicolumn{6}{c}{\textbf{Remove random: Decision flip?}}                                                                                                                                                                                                                  \\
                                          &                           & \multicolumn{2}{c}{Yahoo}                                                                       &  &                           & \multicolumn{2}{c}{IMDB}                                                                        \\
                                          &                           & Yes                                           & No                                           &  &                           & Yes                                           & No                                           \\ \cline{3-4} \cline{7-8} 
                                          & \multicolumn{1}{r|}{Yes} & \multicolumn{1}{c|}{0.7}                         & \multicolumn{1}{c|}{\cellcolor[HTML]{DAE8FC}4.1} &  & \multicolumn{1}{r|}{Yes} & \multicolumn{1}{c|}{0.3}                         & \multicolumn{1}{c|}{\cellcolor[HTML]{DAE8FC}11.1} \\ \cline{3-4} \cline{7-8} 
                                          & \multicolumn{1}{r|}{No} & \multicolumn{1}{c|}{\cellcolor[HTML]{F8A102}1.1} & \multicolumn{1}{c|}{94.1}                         &  & \multicolumn{1}{r|}{No} & \multicolumn{1}{c|}{\cellcolor[HTML]{F8A102}0.4} & \multicolumn{1}{c|}{88.2}                         \\ \cline{3-4} \cline{7-8} 
                                          &                           & \multicolumn{1}{l}{}                           & \multicolumn{1}{l}{}                           &  &                           & \multicolumn{1}{l}{}                           & \multicolumn{1}{l}{}                           \\
\end{tabular}
}

%% file: appendix_stuff/decflip_rand_gradmult_sub_hanrnn.tex
\subfloat[HANrnns]{
\begin{tabular}{clccllcc}
\multicolumn{1}{l}{}                &      & \multicolumn{6}{c}{\textbf{Remove random: Decision flip?}}                                                                                                                                                                                                                  \\
                                          &                           & \multicolumn{2}{c}{Yahoo}                                                                       &  &                           & \multicolumn{2}{c}{IMDB}                                                                        \\
                                          &                           & Yes                                           & No                                           &  &                           & Yes                                           & No                                           \\ \cline{3-4} \cline{7-8} 
                                          & \multicolumn{1}{r|}{Yes} & \multicolumn{1}{c|}{1.3}                         & \multicolumn{1}{c|}{\cellcolor[HTML]{DAE8FC}9.5} &  & \multicolumn{1}{r|}{Yes} & \multicolumn{1}{c|}{2.9}                         & \multicolumn{1}{c|}{\cellcolor[HTML]{DAE8FC}15.5} \\ \cline{3-4} \cline{7-8} 
                                          & \multicolumn{1}{r|}{No} & \multicolumn{1}{c|}{\cellcolor[HTML]{F8A102}0.5} & \multicolumn{1}{c|}{88.7}                         &  & \multicolumn{1}{r|}{No} & \multicolumn{1}{c|}{\cellcolor[HTML]{F8A102}0.6} & \multicolumn{1}{c|}{81.0}                         \\ \cline{3-4} \cline{7-8} 
                                          &                           & \multicolumn{1}{l}{}                           & \multicolumn{1}{l}{}                           &  &                           & \multicolumn{1}{l}{}                           & \multicolumn{1}{l}{}                           \\
\end{tabular}
}

%% file: appendix_stuff/decflip_rand_gradmult_sub_flanrnn.tex
\subfloat[FLANrnns]{
\begin{tabular}{clccllcc}
\multicolumn{1}{l}{}                &      & \multicolumn{6}{c}{\textbf{Remove random: Decision flip?}}                                                                                                                                                                                                                  \\
                                          &                           & \multicolumn{2}{c}{Yahoo}                                                                       &  &                           & \multicolumn{2}{c}{IMDB}                                                                        \\
                                          &                           & Yes                                           & No                                           &  &                           & Yes                                           & No                                           \\ \cline{3-4} \cline{7-8} 
                                          & \multicolumn{1}{r|}{Yes} & \multicolumn{1}{c|}{0.6}                         & \multicolumn{1}{c|}{\cellcolor[HTML]{DAE8FC}13.2} &  & \multicolumn{1}{r|}{Yes} & \multicolumn{1}{c|}{0.2}                         & \multicolumn{1}{c|}{\cellcolor[HTML]{DAE8FC}18.3} \\ \cline{3-4} \cline{7-8} 
                                          & \multicolumn{1}{r|}{No} & \multicolumn{1}{c|}{\cellcolor[HTML]{F8A102}0.0} & \multicolumn{1}{c|}{86.2}                         &  & \multicolumn{1}{r|}{No} & \multicolumn{1}{c|}{\cellcolor[HTML]{F8A102}0.1} & \multicolumn{1}{c|}{81.5}                         \\ \cline{3-4} \cline{7-8} 
                                          &                           & \multicolumn{1}{l}{}                           & \multicolumn{1}{l}{}                           &  &                           & \multicolumn{1}{l}{}                           & \multicolumn{1}{l}{}                           \\
\end{tabular}
}

%% file: appendix_stuff/decflip_rand_gradmult_sub_hanconv.tex
\subfloat[HANconvs]{
\begin{tabular}{clccllcc}
\multicolumn{1}{l}{}                &      & \multicolumn{6}{c}{\textbf{Remove random: Decision flip?}}                                                                                                                                                                                                                  \\
                                          &                           & \multicolumn{2}{c}{Yahoo}                                                                       &  &                           & \multicolumn{2}{c}{IMDB}                                                                        \\
                                          &                           & Yes                                           & No                                           &  &                           & Yes                                           & No                                           \\ \cline{3-4} \cline{7-8} 
                                          & \multicolumn{1}{r|}{Yes} & \multicolumn{1}{c|}{2.8}                         & \multicolumn{1}{c|}{\cellcolor[HTML]{DAE8FC}5.1} &  & \multicolumn{1}{r|}{Yes} & \multicolumn{1}{c|}{5.0}                         & \multicolumn{1}{c|}{\cellcolor[HTML]{DAE8FC}20.5} \\ \cline{3-4} \cline{7-8} 
                                          & \multicolumn{1}{r|}{No} & \multicolumn{1}{c|}{\cellcolor[HTML]{F8A102}0.4} & \multicolumn{1}{c|}{91.7}                         &  & \multicolumn{1}{r|}{No} & \multicolumn{1}{c|}{\cellcolor[HTML]{F8A102}0.9} & \multicolumn{1}{c|}{73.6}                         \\ \cline{3-4} \cline{7-8} 
                                          &                           & \multicolumn{1}{l}{}                           & \multicolumn{1}{l}{}                           &  &                           & \multicolumn{1}{l}{}                           & \multicolumn{1}{l}{}                           \\
\end{tabular}
}

%% file: appendix_stuff/decflip_rand_gradmult_sub_flanconv.tex
\subfloat[FLANconvs]{
\begin{tabular}{clccllcc}
\multicolumn{1}{l}{}                &      & \multicolumn{6}{c}{\textbf{Remove random: Decision flip?}}                                                                                                                                                                                                                  \\
                                          &                           & \multicolumn{2}{c}{Yahoo}                                                                       &  &                           & \multicolumn{2}{c}{IMDB}                                                                        \\
                                          &                           & Yes                                           & No                                           &  &                           & Yes                                           & No                                           \\ \cline{3-4} \cline{7-8} 
                                          & \multicolumn{1}{r|}{Yes} & \multicolumn{1}{c|}{1.1}                         & \multicolumn{1}{c|}{\cellcolor[HTML]{DAE8FC}19.3} &  & \multicolumn{1}{r|}{Yes} & \multicolumn{1}{c|}{0.5}                         & \multicolumn{1}{c|}{\cellcolor[HTML]{DAE8FC}26.7} \\ \cline{3-4} \cline{7-8} 
                                          & \multicolumn{1}{r|}{No} & \multicolumn{1}{c|}{\cellcolor[HTML]{F8A102}0.0} & \multicolumn{1}{c|}{79.6}                         &  & \multicolumn{1}{r|}{No} & \multicolumn{1}{c|}{\cellcolor[HTML]{F8A102}0.0} & \multicolumn{1}{c|}{72.8}                         \\ \cline{3-4} \cline{7-8} 
                                          &                           & \multicolumn{1}{l}{}                           & \multicolumn{1}{l}{}                           &  &                           & \multicolumn{1}{l}{}                           & \multicolumn{1}{l}{}                           \\
\end{tabular}
}

%% file: appendix_stuff/decflip_rand_gradmult_sub_hannoenc.tex
\subfloat[HANnoencs]{
\begin{tabular}{clccllcc}
\multicolumn{1}{l}{}                &      & \multicolumn{6}{c}{\textbf{Remove random: Decision flip?}}                                                                                                                                                                                                                  \\
                                          &                           & \multicolumn{2}{c}{Yahoo}                                                                       &  &                           & \multicolumn{2}{c}{IMDB}                                                                        \\
                                          &                           & Yes                                           & No                                           &  &                           & Yes                                           & No                                           \\ \cline{3-4} \cline{7-8} 
                                          & \multicolumn{1}{r|}{Yes} & \multicolumn{1}{c|}{5.6}                         & \multicolumn{1}{c|}{\cellcolor[HTML]{DAE8FC}23.1} &  & \multicolumn{1}{r|}{Yes} & \multicolumn{1}{c|}{9.2}                         & \multicolumn{1}{c|}{\cellcolor[HTML]{DAE8FC}42.9} \\ \cline{3-4} \cline{7-8} 
                                          & \multicolumn{1}{r|}{No} & \multicolumn{1}{c|}{\cellcolor[HTML]{F8A102}0.6} & \multicolumn{1}{c|}{70.7}                         &  & \multicolumn{1}{r|}{No} & \multicolumn{1}{c|}{\cellcolor[HTML]{F8A102}0.8} & \multicolumn{1}{c|}{47.1}                         \\ \cline{3-4} \cline{7-8} 
                                          &                           & \multicolumn{1}{l}{}                           & \multicolumn{1}{l}{}                           &  &                           & \multicolumn{1}{l}{}                           & \multicolumn{1}{l}{}                           \\
\end{tabular}
}

%% file: appendix_stuff/decflip_rand_gradmult_sub_flannoenc.tex
\subfloat[FLANnoencs]{
\begin{tabular}{clccllcc}
\multicolumn{1}{l}{}                &      & \multicolumn{6}{c}{\textbf{Remove random: Decision flip?}}                                                                                                                                                                                                                  \\
                                          &                           & \multicolumn{2}{c}{Yahoo}                                                                       &  &                           & \multicolumn{2}{c}{IMDB}                                                                        \\
                                          &                           & Yes                                           & No                                           &  &                           & Yes                                           & No                                           \\ \cline{3-4} \cline{7-8} 
                                          & \multicolumn{1}{r|}{Yes} & \multicolumn{1}{c|}{1.8}                         & \multicolumn{1}{c|}{\cellcolor[HTML]{DAE8FC}19.4} &  & \multicolumn{1}{r|}{Yes} & \multicolumn{1}{c|}{0.6}                         & \multicolumn{1}{c|}{\cellcolor[HTML]{DAE8FC}36.2} \\ \cline{3-4} \cline{7-8} 
                                          & \multicolumn{1}{r|}{No} & \multicolumn{1}{c|}{\cellcolor[HTML]{F8A102}0.0} & \multicolumn{1}{c|}{78.8}                         &  & \multicolumn{1}{r|}{No} & \multicolumn{1}{c|}{\cellcolor[HTML]{F8A102}0.1} & \multicolumn{1}{c|}{63.1}                         \\ \cline{3-4} \cline{7-8} 
                                          &                           & \multicolumn{1}{l}{}                           & \multicolumn{1}{l}{}                           &  &                           & \multicolumn{1}{l}{}                           & \multicolumn{1}{l}{}                           \\
\end{tabular}
}

%% file: ms.bbl
\begin{thebibliography}{39}
\expandafter\ifx\csname natexlab\endcsname\relax\def\natexlab#1{#1}\fi

\bibitem[{Adebayo et~al.(2018)Adebayo, Gilmer, Muelly, Goodfellow, Hardt, and
  Kim}]{sanitychecks}
Julius Adebayo, Justin Gilmer, Michael Muelly, Ian Goodfellow, Moritz Hardt,
  and Been Kim. 2018.
\newblock Sanity checks for saliency maps.
\newblock In \emph{Advances in Neural Information Processing Systems}.

\bibitem[{Arras et~al.(2016)Arras, Horn, Montavon, M{\"u}ller, and
  Samek}]{arras2016explaining}
Leila Arras, Franziska Horn, Gr{\'e}goire Montavon, Klaus-Robert M{\"u}ller,
  and Wojciech Samek. 2016.
\newblock Explaining predictions of non-linear classifiers in {N}{L}{P}.
\newblock \emph{arXiv preprint arXiv:1606.07298}.

\bibitem[{Bahdanau et~al.(2015)Bahdanau, Cho, and Bengio}]{bahdanau}
Dzmitry Bahdanau, Kyunghyun Cho, and Yoshua Bengio. 2015.
\newblock Neural machine translation by jointly learning to align and
  translate.
\newblock In \emph{Proceedings of the International Conference on Learning
  Representations}.

\bibitem[{Cheng et~al.(2016)Cheng, Dong, and Lapata}]{cheng2016long}
Jianpeng Cheng, Li~Dong, and Mirella Lapata. 2016.
\newblock Long short-term memory-networks for machine reading.
\newblock In \emph{Proceedings of the Conference on Empirical Methods in
  Natural Language Processing}.

\bibitem[{Diao et~al.(2014)Diao, Qiu, Wu, Smola, Jiang, and Wang}]{imdb}
Qiming Diao, Minghui Qiu, Chao-Yuan Wu, Alexander~J. Smola, Jing Jiang, and
  Chong Wang. 2014.
\newblock Jointly modeling aspects, ratings and sentiments for movie
  recommendation ({J}{M}{A}{R}{S}).
\newblock In \emph{Proceedings of the ACM SIGKDD International Conference on
  Knowledge Ciscovery and Data mining}.

\bibitem[{Ding et~al.(2017)Ding, Liu, Luan, and Sun}]{ding2017visualizing}
Yanzhuo Ding, Yang Liu, Huanbo Luan, and Maosong Sun. 2017.
\newblock Visualizing and understanding neural machine translation.
\newblock In \emph{Proceedings of the 55th Annual Meeting of the Association
  for Computational Linguistics (Volume 1: Long Papers)}.

\bibitem[{Feng et~al.(2018)Feng, Wallace, Grissom~II, Iyyer, Rodriguez, and
  Boyd-Graber}]{pathologies}
Shi Feng, Eric Wallace, Alvin Grissom~II, Mohit Iyyer, Pedro Rodriguez, and
  Jordan Boyd-Graber. 2018.
\newblock Pathologies of neural models make interpretation difficult.
\newblock In \emph{Proceedings of the Conference on Empirical Methods in
  Natural Language Processing}.

\bibitem[{Fyshe et~al.(2015)Fyshe, Wehbe, Talukdar, Murphy, and
  Mitchell}]{fyshe2015compositional}
Alona Fyshe, Leila Wehbe, Partha~P. Talukdar, Brian Murphy, and Tom~M.
  Mitchell. 2015.
\newblock A compositional and interpretable semantic space.
\newblock In \emph{Proceedings of the Conference of the North American Chapter
  of the Association for Computational Linguistics: Human Language
  Technologies}.

\bibitem[{Ghaeini et~al.(2018)Ghaeini, Fern, and
  Tadepalli}]{ghaeini2018interpreting}
Reza Ghaeini, Xiaoli~Z. Fern, and Prasad Tadepalli. 2018.
\newblock Interpreting recurrent and attention-based neural models: a case
  study on natural language inference.
\newblock \emph{arXiv preprint arXiv:1808.03894}.

\bibitem[{Ghorbani et~al.(2017)Ghorbani, Abid, and
  Zou}]{ghorbani2017interpretation}
Amirata Ghorbani, Abubakar Abid, and James Zou. 2017.
\newblock {I}nterpretation of {N}eural {N}etworks is {F}ragile.
\newblock \emph{arXiv preprint arXiv:1710.10547}.

\bibitem[{Habernal et~al.(2018)Habernal, Wachsmuth, Gurevych, and
  Stein}]{beforenamecalling}
Ivan Habernal, Henning Wachsmuth, Iryna Gurevych, and Benno Stein. 2018.
\newblock Before {N}ame-calling: {D}ynamics and {T}riggers of {A}d {H}ominem
  {F}allacies in {W}eb {A}rgumentation.
\newblock \emph{arXiv preprint arXiv:1802.06613}.

\bibitem[{Hermann et~al.(2015)Hermann, Kocisky, Grefenstette, Espeholt, Kay,
  Suleyman, and Blunsom}]{hermann2015teaching}
Karl~Moritz Hermann, Tomas Kocisky, Edward Grefenstette, Lasse Espeholt, Will
  Kay, Mustafa Suleyman, and Phil Blunsom. 2015.
\newblock Teaching machines to read and comprehend.
\newblock In \emph{Advances in Neural Information Processing Systems}.

\bibitem[{Jain and Wallace(2019)}]{attentionexplanation}
Sarthak Jain and Byron~C. Wallace. 2019.
\newblock \href {https://www.aclweb.org/anthology/N19-1357} {{A}ttention is not
  {E}xplanation}.
\newblock In \emph{Proceedings of the 2019 Conference of the North {A}merican
  Chapter of the Association for Computational Linguistics: Human Language
  Technologies, Volume 1 (Long and Short Papers)}.

\bibitem[{Ji and Smith(2017)}]{yangfeng-acl-2017}
Yangfeng Ji and Noah~A. Smith. 2017.
\newblock Neural discourse structure for text categorization.
\newblock In \emph{Proceedings of the Annual Meeting of the Association for
  Computational Linguistics}.

\bibitem[{Kim(2014)}]{kim2014convolutional}
Yoon Kim. 2014.
\newblock Convolutional neural networks for sentence classification.
\newblock In \emph{Proceedings of the Conference on Empirical Methods in
  Natural Language Processing}.

\bibitem[{Kingma and Ba(2014)}]{kingma2014adam}
Diederik~P Kingma and Jimmy Ba. 2014.
\newblock Adam: A method for stochastic optimization.
\newblock \emph{arXiv preprint arXiv:1412.6980}.

\bibitem[{Koehn and Knowles(2017)}]{contextualattn}
Philipp Koehn and Rebecca Knowles. 2017.
\newblock Six challenges for neural machine translation.
\newblock In \emph{Proceedings of the First Workshop on Neural Machine
  Translation}.

\bibitem[{Koh and Liang(2017)}]{influencefns}
Pang~Wei Koh and Percy Liang. 2017.
\newblock Understanding black-box predictions via influence functions.
\newblock \emph{arXiv preprint arXiv:1703.04730}.

\bibitem[{Lee et~al.(2017)Lee, Shin, and Kim}]{lee2017interactive}
Jaesong Lee, Joong-Hwi Shin, and Jun-Seok Kim. 2017.
\newblock {I}nteractive {V}isualization and {M}anipulation of {A}ttention-based
  {N}eural {M}achine {T}ranslation.
\newblock In \emph{Proceedings of the 2017 Conference on Empirical Methods in
  Natural Language Processing: System Demonstrations}, pages 121--126.

\bibitem[{Lei et~al.(2016)Lei, Barzilay, and Jaakkola}]{lei2016rationalizing}
Tao Lei, Regina Barzilay, and Tommi Jaakkola. 2016.
\newblock Rationalizing neural predictions.
\newblock In \emph{Proceedings of the Conference on Empirical Methods in
  Natural Language Processing}.

\bibitem[{Li et~al.(2015)Li, Chen, Hovy, and Jurafsky}]{compositionality}
Jiwei Li, Xinlei Chen, Eduard Hovy, and Dan Jurafsky. 2015.
\newblock Visualizing and understanding neural models in {N}{L}{P}.
\newblock \emph{arXiv preprint arXiv:1506.01066}.

\bibitem[{Li et~al.(2016)Li, Monroe, and Jurafsky}]{li2016understanding}
Jiwei Li, Will Monroe, and Dan Jurafsky. 2016.
\newblock Understanding neural networks through representation erasure.
\newblock \emph{arXiv preprint arXiv:1612.08220}.

\bibitem[{Lin and Bilmes(2011)}]{lin2011class}
Hui Lin and Jeff Bilmes. 2011.
\newblock A class of submodular functions for document summarization.
\newblock In \emph{Proceedings of the Annual Meeting of the Association for
  Computational Linguistics: Human Language Technologies-Volume 1}.

\bibitem[{Lin et~al.(2017)Lin, Feng, Santos, Yu, Xiang, Zhou, and
  Bengio}]{2dsentemb}
Zhouhan Lin, Minwei Feng, Cicero Nogueira~dos Santos, Mo~Yu, Bing Xiang, Bowen
  Zhou, and Yoshua Bengio. 2017.
\newblock A structured self-attentive sentence embedding.
\newblock \emph{arXiv preprint arXiv:1703.03130}.

\bibitem[{Lipton(2016)}]{mythos}
Zachary~C Lipton. 2016.
\newblock The mythos of model interpretability.
\newblock \emph{arXiv preprint arXiv:1606.03490}.

\bibitem[{Liu and Lapata(2018)}]{liu2018learning}
Yang Liu and Mirella Lapata. 2018.
\newblock Learning structured text representations.
\newblock \emph{Transactions of the Association of Computational Linguistics},
  6:63--75.

\bibitem[{Luong et~al.(2015)Luong, Pham, and Manning}]{luong2015effective}
Minh-Thang Luong, Hieu Pham, and Christopher~D. Manning. 2015.
\newblock Effective approaches to attention-based neural machine translation.
\newblock \emph{arXiv preprint arXiv:1508.04025}.

\bibitem[{Martins and Astudillo(2016)}]{sparsemax}
Andr\'e Martins and Ram\'on Astudillo. 2016.
\newblock {F}rom softmax to sparsemax: {A} sparse model of attention and
  multi-label classification.
\newblock In \emph{International Conference on Machine Learning}.

\bibitem[{Melis and Jaakkola(2018)}]{melis2018towards}
David~Alvarez Melis and Tommi Jaakkola. 2018.
\newblock Towards robust interpretability with self-explaining neural networks.
\newblock In \emph{Advances in Neural Information Processing Systems}.

\bibitem[{Nguyen(2018)}]{nguyen}
Dong Nguyen. 2018.
\newblock {C}omparing automatic and human evaluation of local explanations for
  text classification.
\newblock In \emph{Proceedings of the Conference of the North American Chapter
  of the Association for Computational Linguistics: Human Language
  Technologies, Volume 1 (Long Papers)}.

\bibitem[{Ribeiro et~al.(2016)Ribeiro, Singh, and Guestrin}]{lime}
Marco~Tulio Ribeiro, Sameer Singh, and Carlos Guestrin. 2016.
\newblock ``{W}hy should {I} trust you?'': Explaining the predictions of any
  classifier.
\newblock In \emph{Proceedings of the ACM SIGKDD International Conference on
  Knowledge Discovery and Data mining}.

\bibitem[{Ross et~al.(2017)Ross, Hughes, and Doshi-Velez}]{rightreasons}
Andrew~Slavin Ross, Michael~C. Hughes, and Finale Doshi-Velez. 2017.
\newblock Right for the right reasons: Training differentiable models by
  constraining their explanations.
\newblock In \emph{Proceedings of the International Joint Conference on
  Artificial Intelligence}.

\bibitem[{Subramanian et~al.(2017)Subramanian, Pruthi, Jhamtani,
  Berg-Kirkpatrick, and Hovy}]{spine}
Anant Subramanian, Danish Pruthi, Harsh Jhamtani, Taylor Berg-Kirkpatrick, and
  Eduard Hovy. 2017.
\newblock {S}{P}{I}{N}{E}: Sparse interpretable neural embeddings.
\newblock \emph{arXiv preprint arXiv:1711.08792}.

\bibitem[{Vaswani et~al.(2017)Vaswani, Shazeer, Parmar, Uszkoreit, Jones,
  Gomez, Kaiser, and Polosukhin}]{transformer}
Ashish Vaswani, Noam Shazeer, Niki Parmar, Jakob Uszkoreit, Llion Jones,
  Aidan~N. Gomez, {\L}ukasz Kaiser, and Illia Polosukhin. 2017.
\newblock Attention is all you need.
\newblock In \emph{Advances in Neural Information Processing Systems}.

\bibitem[{Vinyals et~al.(2015)Vinyals, Kaiser, Koo, Petrov, Sutskever, and
  Hinton}]{vinyals2015grammar}
Oriol Vinyals, {\L}ukasz Kaiser, Terry Koo, Slav Petrov, Ilya Sutskever, and
  Geoffrey Hinton. 2015.
\newblock Grammar as a foreign language.
\newblock In \emph{Advances in Neural Information Processing Systems}.

\bibitem[{Wang et~al.(2016)Wang, Huang, Zhao et~al.}]{wang2016attention}
Yequan Wang, Minlie Huang, Li~Zhao, et~al. 2016.
\newblock Attention-based {L}{S}{T}{M} for aspect-level sentiment
  classification.
\newblock In \emph{Proceedings of the 2016 Conference on Empirical Methods in
  Natural Language Processing}, pages 606--615.

\bibitem[{Yang et~al.(2017)Yang, Mukherjee, and Dragut}]{yang2017satirical}
Fan Yang, Arjun Mukherjee, and Eduard Dragut. 2017.
\newblock Satirical news detection and analysis using attention mechanism and
  linguistic features.
\newblock \emph{arXiv preprint arXiv:1709.01189}.

\bibitem[{Yang et~al.(2016)Yang, Yang, Dyer, He, Smola, and Hovy}]{han}
Zichao Yang, Diyi Yang, Chris Dyer, Xiaodong He, Alex Smola, and Eduard Hovy.
  2016.
\newblock Hierarchical attention networks for document classification.
\newblock In \emph{Proceedings of the Conference of the North American Chapter
  of the Association for Computational Linguistics: Human Language
  Technologies}.

\bibitem[{Zhang et~al.(2015)Zhang, Zhao, and LeCun}]{yahooamazon}
Xiang Zhang, Junbo Zhao, and Yann LeCun. 2015.
\newblock Character-level convolutional networks for text classification.
\newblock In \emph{Advances in Neural Information Processing Systems}.

\end{thebibliography}
